# Origami spring-inspired shape morphing for flexible robotics

*Qianying Chen*[1,2], *Fan Feng*[1], *Pengyu Lv*[1], *Huiling Duan*[1,2*]

[1]State Key Laboratory for Turbulence and Complex Systems, Department of Mechanics and Engineering Science, BIC-ESAT, College of Engineering, Peking University, Beijing, China

[2]CAPT, HEDPS and IFSA Collaborative Innovation Center of MoE, Peking University, Beijing, China

[*]Correspondence should be addressed to H.D. (email:hlduan@pku.edu.cn)

## Abstract

Flexible robotics are capable of achieving various functionalities by shape morphing, benefiting from their compliant bodies and reconfigurable structures. Here we construct and study a class of origami springs generalized from the known interleaved origami spring, as promising candidates for shape morphing in flexible robotics. These springs are found to exhibit nonlinear stretch-twist coupling and linear/nonlinear mechanical response in the compression/tension region, analyzed by the demonstrated continuum mechanics models, experiments, and finite element simulations. To improve the mechanical performance such as the damage resistance, we establish an origami rigidization method by adding additional creases to the spring system. Guided by the theoretical framework, we experimentally realize three types of flexible robotics — origami spring ejectors, crawlers, and transformers. These robots show the desired functionality and outstanding mechanical performance. The proposed concept of origami-aided design is expected to pave the way to facilitate the diverse shape morphing of flexible robotics.

## Introduction

Flexible robotics possess compliant bodies and reconfigurable structures to achieve multiple functionalities, such as delivering different objects,[1,2] and going through diverse environments.[3,4] The essential performance for functionality is the controllable shape morphing, based on materials or structural design or both. The material-based shape morphing can be classified into passive shape morphing with soft materials and active shape morphing with active materials. The emphasis of the former is to seek efficient actuation strategies like pneumatic, hydraulic, or chemical energy to drive the shape changing of robots,[1,4,5] while the latter is to explore advanced active materials to realize specific deformations in response to the external stimuli.[6-13]

Shape-morphing strategies of flexible robotics based on structural design are focused on exploiting special structures with specific mechanical properties, such as negative Poisson's ratio,[14,15] multistability,[16-18] and buckling[19]. These structures can provide various and programmable deformation modes, and thus have many advantages and great potential prospects.[20] Among various structural design strategies, origami made of thin paper is a promising candidate.[2,21-27] First, origami is constructed by folding the paper along the prescribed creases on an easy-to-manufacture two-dimensional sheet. The distribution of creases can be designed to provide desired geometrical constraints, leading to different degrees of freedom of motion and potentially simple control systems.[23,28-30] Second, the bending effect usually dominates in the origami system, introducing generically nonlinear effective mechanical response, which can be utilized to design origami springs,[24,28,31] origami cantilevers,[32,33] etc. Despite having various designs of origami actuators/robotics in the literature, these applications usually suffer from difficulties: 1. The models to capture the shape morphing rules and the mechanical response are usually sophisticated (e.g., computationally expensive truss model[34,35]), which enhance the difficulties of the rational design; 2. There usually exists inevitable stress concentration at the corner of the crease (e.g., Kresling pattern[22]), which results in undesired damage to the structure. Therefore, it is necessary to develop new theoretical methods to overcome these difficulties and guide the further use of origami in flexible robotics.

In this work, we consider the generalized origami springs as a template for the application of flexible robotics, which also overcomes the above-mentioned difficulties to some extent. To this end, we firstly introduce a series of origami springs by alternately folding multiple regular-polygon-array paper ribbons, known as the interleaved origami spring (IOS). IOSs exhibit stretch-twist-coupling shape morphing when stretching. We demonstrate a continuum model to simplify the analysis of the shape morphing and the mechanical response of IOSs. The theoretical model is validated by experiments and finite element simulations. Moreover, to optimize the mechanical response and improve the damage resistance, we propose an origami rigidization method by predesigning additional creases on the facets to transform the buckled facets into rigidly foldable facets. The simulation shows that the origami rigidization significantly lowers the stress concentration and enhances the damage resistance of the structure. Then, the shape-morphing performances and mechanical responses of IOSs are further applied to construct flexible robotics, including origami spring ejectors, crawlers, and transformers, which demonstrate the feasibility of the origami-aided-design concept. Finally, we anticipate that the design framework can be generalized and applied to other systems such as stimuli-sensitive and non-periodic origami.

## Construction of the origami spring

The zigzag origami (ZO) is the simplest origami structure to provide elasticity of a rectangular paper ribbon through the alternate mountain-valley folding. Applying the uniaxial force $F$, the ZO behaves as a nonlinear spring with no twist involved (Fig. 1a). Inspired by the ZO, we construct an interleaved origami spring (IOS) by alternately folding two perpendicularly arranged rectangular paper ribbons of the same size (Fig. 1b). In addition to elasticity, the IOS also exhibits a coupled twist along the central axis (Fig. 1c) when stretching, due to the interactions between these two paper ribbons. To enrich the class of IOS, we systematically fold ribbons consisting of chains of different polygons. These IOSs are named as IOS-6, and IOS-8 respectively, depending on the shapes of the facets (see Fig. 2 and Supplementary Note S1). They are expected to have distinct stretch-twist couplings and mechanical responses due to the different interactions of facets, which will be analyzed later.

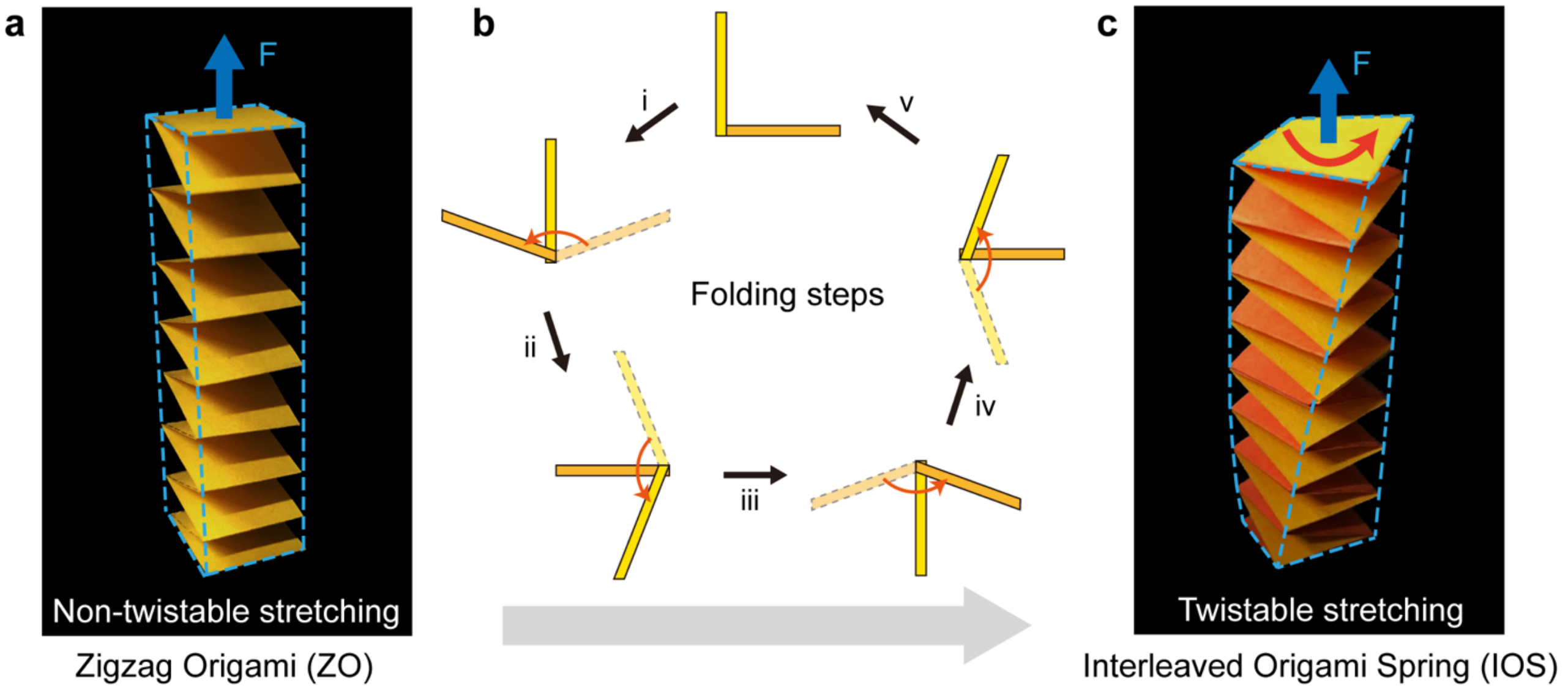

**FIG. 1.** Geometry of the zigzag origami (ZO) and interleaved origami spring (IOS). **(a)** The uniform-stretching configuration of ZO under uniaxial tension. **(b)** The folding process of IOS-4. **(c)** The stretch-twist-coupling shape morphing of IOS under uniaxial tension.

## Shape morphing modeling

In this section, we aim to analyze the stretch-twist coupling of IOSs, by replacing the origami structure with an analogous continuum having the same effective response. This method makes the shape-morphing analysis easier and more intuitive, and also can be employed for other kinds of origami structures. Fig. 2a shows the IOS-4 configurations and the corresponding analogous continuum diagrams in different extension ratios with eight unit cells. The continuum consists of a cylinder with four helices winded on the surface. Three assumptions are satisfied in the simplified model. First, all the sides keep straight with a constant side length $a$; second, the process is quasi-static, while all the unit cells keep the same configuration; third, all the origami vertices land on the corresponding helices. The stretch-twist performance is characterized by two dimensionless parameters, i.e., the twisting angle of each cell $\tilde{\theta}$, and the extension ratio $\tilde{z}$, which are defined as

$$\tilde{\theta} = \frac{\theta}{n} \tag{1}$$

$$\tilde{z} = \frac{z}{l_{\mathrm{LPH}}(a, n, z)} = \sin\varphi \tag{2}$$

where $\theta$ and $z$ are the twisting angle and the extension length of the whole structure, respectively. $n$ is the number of the unit cells, $l_{\mathrm{LPH}}(a, n, z)$ is the arc length of the large-pitch helix, which is a function of $a$, $n$, and $z$, and $\varphi$ is its helix angle.

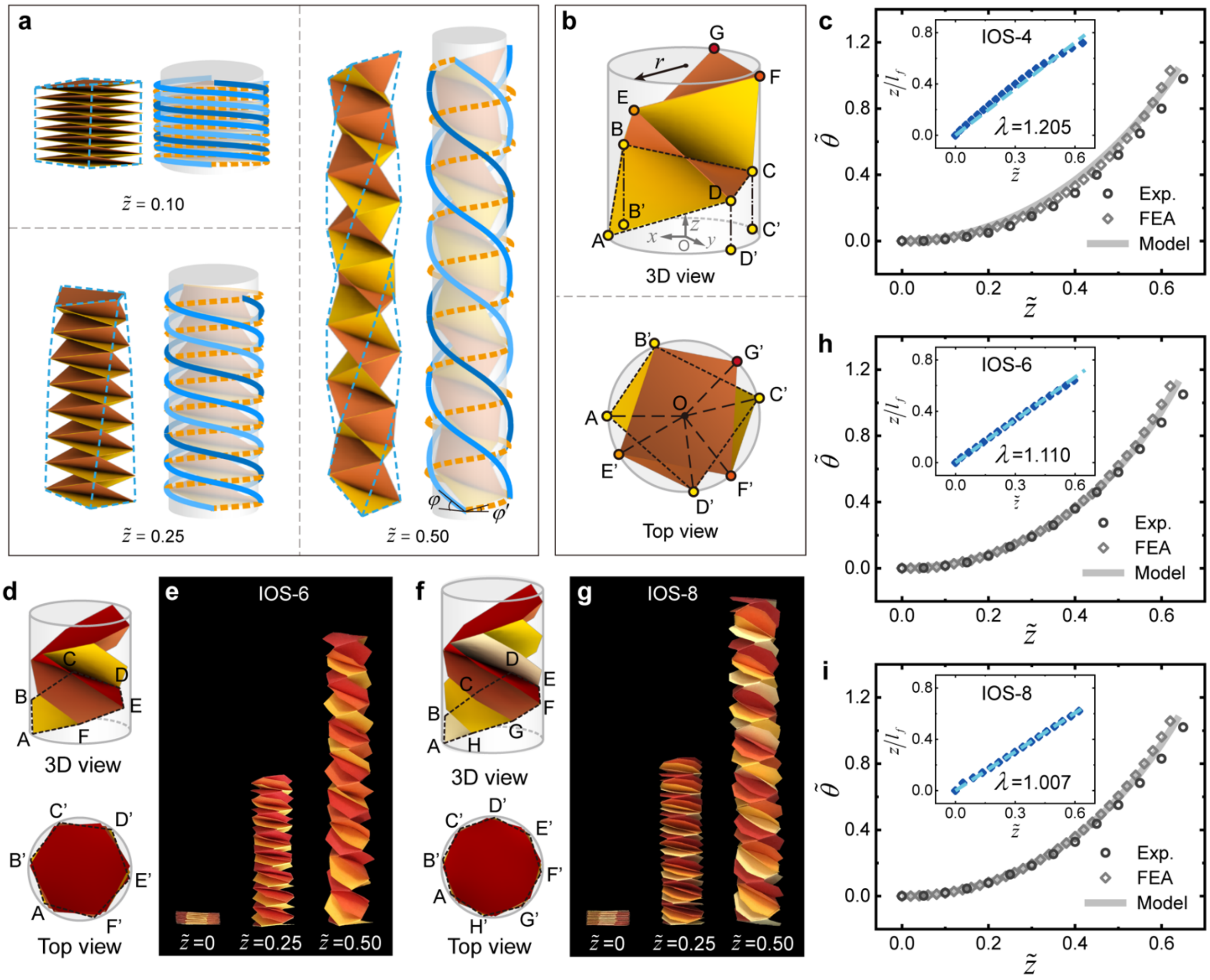

**FIG. 2.** Schematics of the stretch-twist-coupling performance of IOSs and continuum analysis diagrams. **(a)** The configurations of IOS-4 and the corresponding simplified equivalent continuum diagrams in extension ratio $\tilde{z}$ =0.10, 0.25, and 0.50, respectively. Light-, sky-, and dark-blue solid lines indicate the large-pitch helices, while yellow dashed lines the small-pitch. **(b)** The shape-morphing model of the IOS-4 unit cell on 3D and top view, respectively. **(c)** The dependences of $\tilde{\theta}$ upon $\tilde{z}$ for IOS-4. The data were obtained from experiments, FEA, and the proposed theoretical model. The thumbnail exhibits $z/l_f$ as a function of $\tilde{z}$, showing an approximately linear relationship. **(d-i)** Similar analyses for IOS-6 and IOS-8. The results reveal the different stretch-twist couplings in IOS-4, IOS-6, and IOS-8.

We intercept four adjacent facets of an IOS-4 as the unit cell, which contains two square facets in each paper ribbon (Fig. 2b). Projecting the vertices of the unit cell onto the plane perpendicular to the cylindrical axis and passing through A, one can obtain the projected vertices B',...,G' and the sum of the projected arcs $\widehat{AB'} + \widehat{B'C'} + \widehat{C'D'} + \widehat{D'A} = 2\pi r$, where $r$ is the radius of the cylinder. The expression of $r$ is further deduced as a function of $\tilde{z}$ (see Supplementary Note S2 for the details)

$$r = \frac{aN_4}{6}\sqrt{\frac{N_4(N_4 + M_4)}{6}} \quad (3)$$

by calculating the projected arcs, where $M_4 = \sqrt{1 - \tilde{z}^2}$, $N_4 = \sqrt{9 - \tilde{z}^2}$. The twisting angle of each unit cell $\tilde{\theta}$ is also given as a function of $\tilde{z}$

$$\tilde{\theta}(\tilde{z}) = \frac{4(\widehat{E'B'} - \widehat{AB'})}{r} = 4\arccos(P_4 Q_4 + \sqrt{1 - P_4^2}\sqrt{1 - Q_4^2}) \quad (4)$$

where $P_4 = 1 - 108M_4^2/[N_4^3(M_4 + N_4)]$, $Q_4 = 1 - 12/[N_4(M_4 + N_4)]$. We obtain the twisting angle $\theta$ and the corresponding extension length $z$ through the finite element analysis (FEA) and experiments. $\tilde{\theta}$ and $\tilde{z}$ can be further calculated by Eq. (1) and Eq. (4), respectively. The relationship between $z$ and $\tilde{z}$ is approximately linear (Fig. 2c, inset), implying that $l_{\mathrm{LPH}}(a, n, z)$ can be regarded as a constant independent of $z$. We thus assume

$$\tilde{z} = \frac{z}{\lambda l_{\mathrm{f}}}, \lambda \geq 1 \quad (5)$$

where $\lambda l_{\mathrm{f}} = l_{\mathrm{LPH}}$, $\lambda$ is a constant coefficient and $l_{\mathrm{f}}$ is the extension length of the IOS when $\varphi = \pi/2$ ($l_{\mathrm{f}} = 2an$ in IOS-4s). In Fig. 2c, the dependences of $\tilde{\theta}$ upon $\tilde{z}$ (i.e., the stretch-twist coupling) from experiments, FEA, and our proposed model are plotted, which are in great agreements.

Similar analyses are also performed for IOS-6 and IOS-8. The geometric parameters of IOSs with differently shaped facets but the same circumradius $r_0$ are listed in Supplementary Table S1 (derivation process see Supplementary Note S2). The theoretical, FEA and experimental results in Figs. 2h and i reveal that similar stretch-twist couplings also exist in IOS-6 and IOS-8, with slight discrepancies. It is noted that $\lambda$ tends to 1 when increasing the number of the regular-polygon edges, which means, as the number of the edges increases, the length of the large-pitch helix is close to the extension length when $\varphi = \pi/2$.

**Structural optimization**

We observe that the IOSs undergo more and more significant buckling with the progress of stretching in both experiments and simulations. The buckling results in the increase of elastic energy, and hence the increase of tensile force. Moreover, stress concentrations mostly appear near the corners of creases (see Fig. 3b) and cause undesired damages. To resolve these, we propose a predesign-crease strategy, i.e., setting additional creases (namely secondary creases, red lines in Fig. 3a) along the diagonals of facets in addition to the primary creases (black lines in Fig. 3a) along the sides to guide the predictable folding rather than buckling. Similar buckling-control strategies were reported but with different purposes and effects.[29,38-40] By setting these secondary creases, the overwhelmed buckling is released and the buckling conflicts between adjacent facets are relieved, resulting in great improvements of mechanical performances. This strategy essentially transforms the buckled facets into rigidly foldable facets, and therefore is named *origami rigidization method*. Accordingly, the abbreviated name of the IOS optimized by this method is thus updated as RIOS.

Fig. 3b shows the FEA results of the shape-morphing configurations of IOSs and RIOSs ($r_0 = 10$ mm, $n = 8$) colored by the Mises stress nephogram. One can observe that RIOSs exhibit rigidity or approximate rigidity on the facets without or with less buckling, compared to the corresponding IOS cases, especially in the square-facet cases (IOS-4 and RIOS-4). More importantly, the values of the maximal stress decrease in RIOSs compared to those in IOSs, indicating the stress concentrations are relieved.

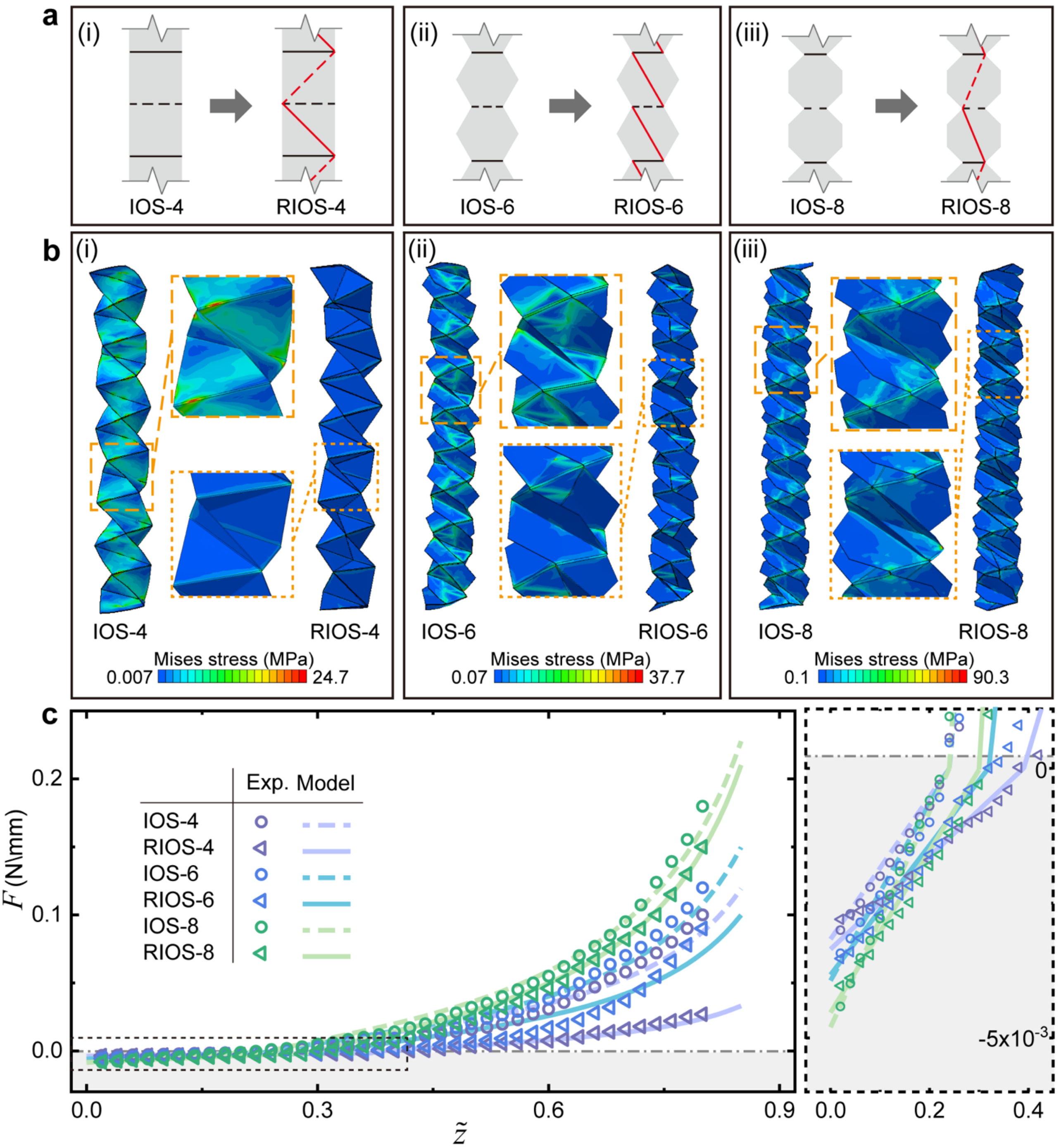

**FIG. 3.** Demonstration of the origami rigidization method. **(a)** Diagrams of the crease distributions. Black (red) lines represent primary (secondary) creases. Solid (dashed) lines are mountain (valley) creases. **(b)** Shape-morphing configurations ($\tilde{z}$ =0.8) obtained by FEA, colored by the Mises stress nephogram. **(c)** Uniaxial force $\tilde{F}$ applied on origami springs versus extension ratio $\tilde{z}$. The theoretical models match the experimental data well.

Next, we analyze the mechanical response of IOSs and RIOSs. Fig. 3c illustrates the force-extension curves of IOSs and RIOSs. $\tilde{F}$ is defined as $\tilde{F} = F/(2ann_{\mathrm{pr}})$, where $2ann_{\mathrm{pr}}$ is the total length of the primary creases with $n_{\mathrm{pr}}$ the number of paper ribbons. Overall, the forces required to stretch the springs follow IOS-4<IOS-6<IOS-8, RIOS-4<RIOS-6<RIOS-8, and RIOS<IOS under the same facet shape, while the square-facet case has the maximal difference. The force increases monotonically and continuously with respect to the uniaxial strain, which indicates no multistability or other special mechanical responses occur. Since all the origami springs exhibit an initial free extension $\tilde{z}_0$ due to the residual strain in creases, we separate the force-extension curves into compression and tension regions.

Observed from the compression region, $\tilde{F}$ and $\tilde{z}$ are in an approximately linear relationship, we thus assume

$$\tilde{F} = k_{\mathrm{c}}(\tilde{z} - \tilde{z}_0) \tag{6}$$

where $k_{\mathrm{c}}$ is the compression elastic constant with the unit $\mathrm{N} \cdot \mathrm{mm}^{-1}$ and can be obtained by linear fitting with the experimental data. As illustrated in Fig. 3c right, the regular-octagon-facet origami springs have the maximal $k_{\mathrm{c}}$ both in IOS and RIOS. Moreover, RIOSs have smaller $k_{\mathrm{c}}$ than IOSs, which illustrates that the rigidization optimization on RIOSs reduces the interaction among the facets to a certain extent.

In the tension region, the force-extension relationship is no longer linear (see Fig. 3c left), which could be attributed to the stronger and stronger interactions between the adjacent facets as the extension progresses. The total shape-morphing energy $E$ is given by the sum of the folding energy of the primary creases $E_{\mathrm{F1}}$, the folding energy of the secondary creases $E_{\mathrm{F2}}$ ($E_{\mathrm{F2}} = 0$ in IOSs), and the buckling energy of the facets $E_{\mathrm{B}}$ ($E_{\mathrm{B}} = 0$ in RIOSs) with the forms

$$\begin{cases} E_{\mathrm{F1}} = \dfrac{1}{2} \cdot f_{\mathrm{k_F}}(\tilde{z}) \cdot 2ann_{\mathrm{pr}} \cdot \omega_1^2 \\ E_{\mathrm{B}} = \dfrac{1}{2} \cdot f_{\mathrm{k_B}}(\tilde{z}) \cdot 4r_0 nn_{\mathrm{pr}} \cdot 2\omega_{\mathrm{B}}^2 \\ E_{\mathrm{F2}} = \dfrac{1}{2} \cdot f_{\mathrm{k_F}}(\tilde{z}) \cdot 4r_0 nn_{\mathrm{pr}} \cdot \omega_2^2 \end{cases} \tag{7}$$

where $f_{\mathrm{k_F}}(\tilde{z})$ and $f_{\mathrm{k_B}}(\tilde{z})$ are the unknown stiffness of the folding creases and the buckling, respectively. Here we assume $f_{\mathrm{k_F}}(\tilde{z}) = k_{\mathrm{F}}\tilde{z}^{\xi_{\mathrm{F}}}$ and $f_{\mathrm{k_B}}(\tilde{z}) = k_{\mathrm{B}}\tilde{z}^{\xi_{\mathrm{B}}}$ to fit the experimental data, where $k_{\mathrm{F}}$ and $k_{\mathrm{B}}$ are the stiffness constants with the unit $\mathrm{N} \cdot \mathrm{rad}^{-1}$, $\xi_{\mathrm{F}}$ and $\xi_{\mathrm{B}}$ are dimensionless constant exponentials in order to capture the nonlinear behavior of $\tilde{F}$ upon $\tilde{z}$. The folding angle of the primary crease $\omega_1$ can be approximately expressed as $\omega_1 = \varphi + \varphi'$ (Supplementary Figure S3). The buckling of the facets is simplified by two equal folding angles $\omega_\mathrm{B}$ along the diagonals, and the folding angle of the secondary creases is denoted by $\omega_2$. All the folding angles are functions of $\tilde{z}$, obtained by evaluating coordinates (see Supplementary Note S2). We use $\tilde{E} = E/2ann_\mathrm{pr}$ to denote the energy per unit length of the primary crease, therefore $\tilde{F}$ can be expressed as the partial derivative of $\tilde{E}$ with respect to $\tilde{z}$, more specifically

$$\begin{cases} \tilde{F}_\mathrm{IOS} = \dfrac{\partial(\tilde{E}_\mathrm{F1} + \tilde{E}_\mathrm{B})}{\partial z} = \dfrac{1}{\lambda l_\mathrm{f}|_{n=1}}[\tilde{z}^{\xi_\mathrm{F}} k_\mathrm{F}(A_1 + A_2\xi_\mathrm{F}) + \tilde{z}^{\xi_\mathrm{B}} k_\mathrm{B}(B_1 + B_2\xi_\mathrm{B})] \\ \tilde{F}_\mathrm{RIOS} = \dfrac{\partial(\tilde{E}_\mathrm{F1} + \tilde{E}_\mathrm{F2})}{\partial z} = \dfrac{1}{\lambda l_\mathrm{f}|_{n=1}}[\tilde{z}^{\xi_\mathrm{F}} k_\mathrm{F}(A_1 + A_2\xi_\mathrm{F} + C_1 + C_2\xi_\mathrm{F})] \end{cases} \quad (8)$$

where $A_1$, $A_2$, $B_1$, $B_2$, $C_1$, and $C_2$ are the geometric correlation parameters related to $\tilde{z}$. $\xi_\mathrm{F}$, $\xi_\mathrm{B}$, $k_\mathrm{F}$, and $k_\mathrm{B}$ are undetermined coefficients. Substituting $\tilde{z} = \tilde{z}_0$, $\tilde{F}_\mathrm{IOS} = 0$ and $\tilde{F}_\mathrm{RIOS} = 0$ into Eq. (8), we have $(\xi_\mathrm{F})_\mathrm{IOS} = -A_1(\tilde{z}_0)/A_2(\tilde{z}_0)$, $(\xi_\mathrm{B})_\mathrm{IOS} = -B_1(\tilde{z}_0)/B_2(\tilde{z}_0)$, and $(\xi_\mathrm{F})_\mathrm{RIOS} = -(A_1(\tilde{z}_0) + C_1(\tilde{z}_0))/(A_2(\tilde{z}_0) + C_2(\tilde{z}_0))$. Then the stiffness constants $k_\mathrm{F}$ and $k_\mathrm{B}$ can be further obtained by fitting with the experimental data using the least square method. Specific values and expressions above are listed in Supplementary Table S2. As shown in Fig. 3c, the force-extension model we proposed matches well with the experimental data with acceptable minor deviations, indicating that the proposed models (Eq. (6) and Eq. (8)) are able to capture the linear and nonlinear mechanical responses in the compression and tension regions. Moreover, the force-displacement curve of RIOS-8 (and IOS-8) has the biggest slope compared with the other types of RIOS (and IOS). This fact can be explained theoretically by Eq. (7) and Eq. (8) – RIOS-8 has the longest crease per unit cell, and hence the largest elastic energy density.

**Performance in flexible robotics**

*1. Origami spring ejector.*

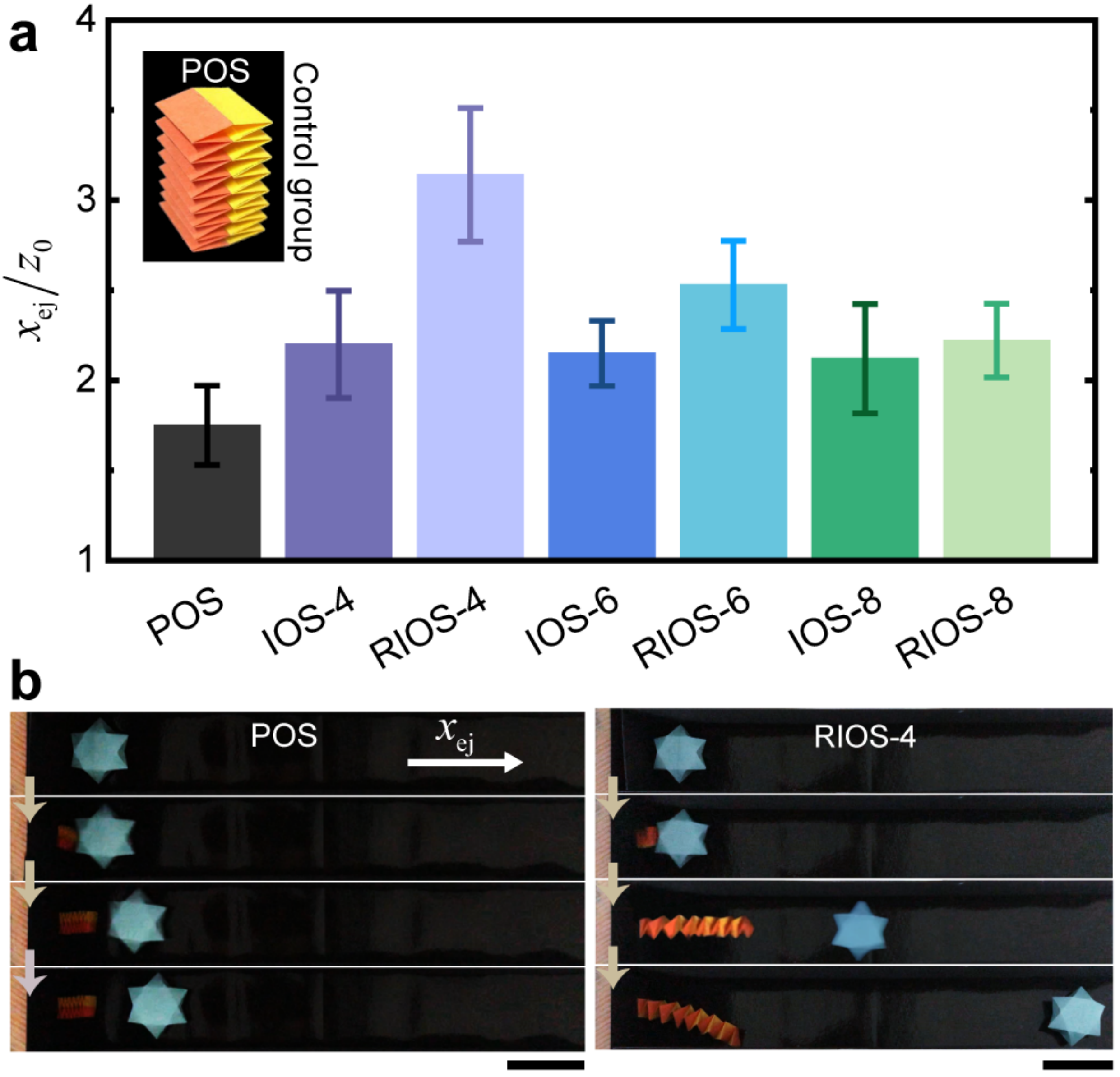

**FIG. 4.** Performance of origami spring ejectors. **(a)** The ejecting performance $x_{ej}/z_0$ of different origami springs. $x_{ej}$ and $z_0$ are the ejecting distance of the origami hexagram and the initial length of the origami spring, respectively. **(b)** The ejecting process of POS and RIOS-4. All involved origami springs are under $r_0$ =10 mm and $n$ =8. Scale bar: 50 mm.

Since the origami springs have excellent compressing elasticity, they can be employed as ejectors. We arrange two ZOs in parallel with the same size and number of creases as IOS-4, named parallel origami spring (POS), to be the control group. All origami springs were firstly compressed to the minimal length and rapidly released to push the origami hexagram for a distance ejecting. Several factors determine the ejecting performance, including $z_0$ and $k_c$ mentioned in the last section, the friction coefficient $\mu$ with the substrate, the mass of the origami spring $M$ and the hexagram $m$ (Supplementary Table S3). The ejecting distance can be deduced by the conservation of energy

$$x_{\mathrm{ej}} = \frac{z_0(k_{\mathrm{c}}z_0 - \mu Mg)}{2\mu mg} \tag{9}$$

where $g$ is the gravitational acceleration constant. Here, we ignore the small reciprocating motion of the origami spring for simplification. As seen from Eq. (9), it requires to increase the complete compression force of origami springs, i.e. $k_{\mathrm{c}}z_0$ and decrease $M$ to obtain a good ejection ability. Owing to the origami rigidization method, the $k_{\mathrm{c}}z_0$ of RIOSs improves to a certain extent, leading to a better ejecting ability when compared with IOSs and POS (Fig. 4a). Among all, the RIOS-4 shows the best performance, which can push the origami hexagram forward over three times longer than its original length (Fig. 4b, Supplementary Movie S1).

*2. Origami spring crawler.*

The origami spring crawler employs the outstanding recovery ability of the spring from its extended state to the free state. The elastic recovery performance was tested by repeatedly stretching the origami springs from their free extension states to $\tilde{z} = 0.8$, and recording the free extensions after unloading. Benefited from the mutual restraints and interactions between the paper ribbons, IOSs and RIOSs have both greatly better recovery performance than POS (Fig. 5a). Therefore, we employed the IOS-8, which had the best elastic recoverability among, to mimic the crawl locomotion of earthworms. The two ends of the origami springs were dragged by rotary motors through two thin threads. The two threads stayed along the structure's central axis to obtain a controllable straight-line crawling. Firstly, we made one of the motor work to stretch the origami spring until $\tilde{z} = 0.8$. Then we stopped that motor and started the other motor to recover the stretching by relaxing the thread. The origami spring contracted to its free extension state with the head as the fixed point, and finished an entire crawling cycle (Fig. 5b, Supplementary Movie S2).

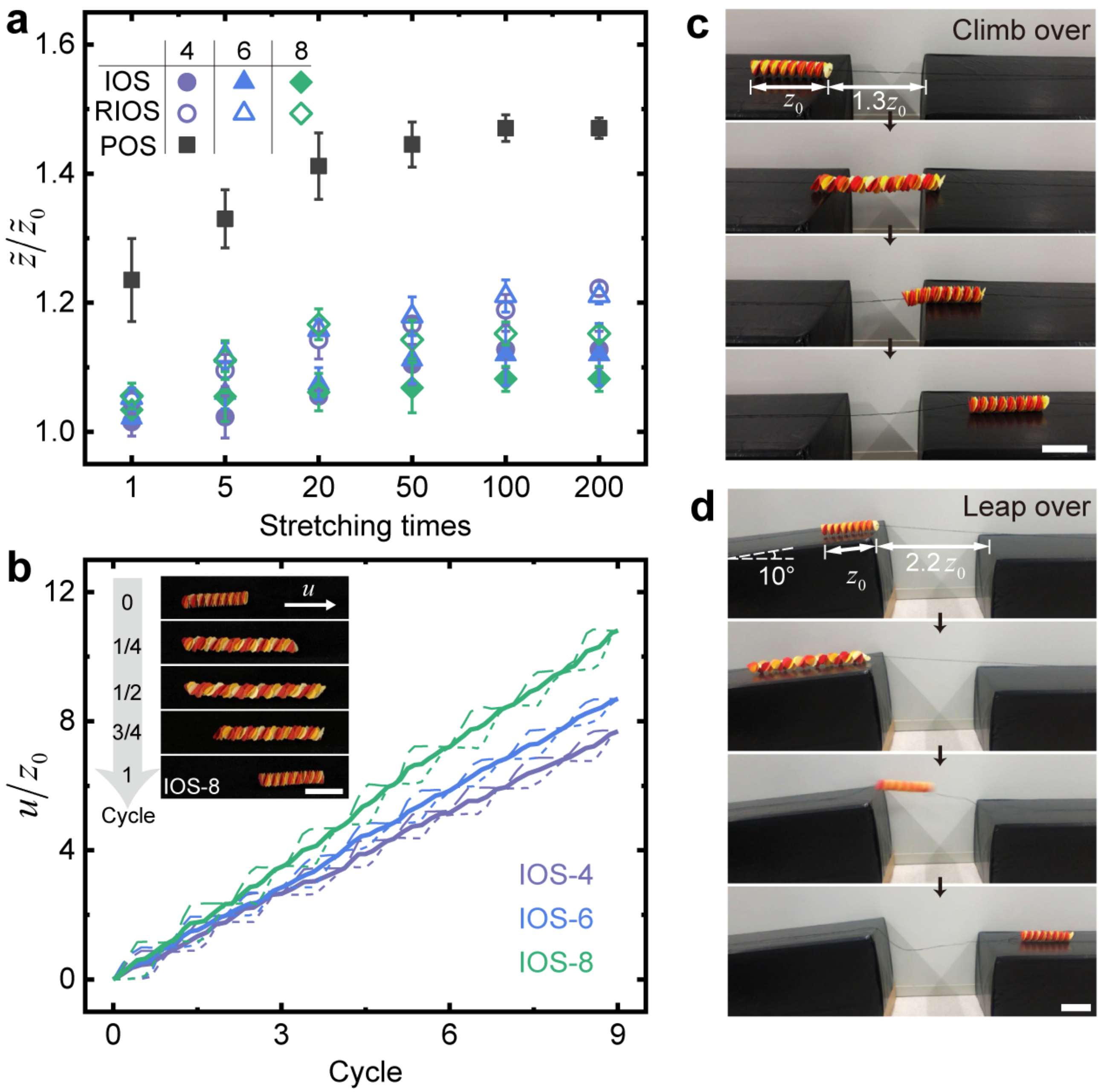

**FIG. 5.** Performance of origami spring crawlers. **(a)** The elastic recovery performance of different origami springs under multiple stretching times. **(b)** The crawling performance $u/z_0$ ($u$ is the crawling distance) of the center of body (solid line), head (dashed line), and tail (dotted line) varying with number of crawling cycles. The thumbnail is five typical crawling moments of the IOS-8 throughout an entire crawling cycle. Crawling images of IOS-4 and IOS-6 can be seen in Supplementary Figure S12. **(c)** The process of the IOS-8 climbing over a cliff with 1.3-fold body length. **(d)** The process of the IOS-8 leaping over a cliff with 2.2-fold body length. All involved origami springs are under $r_0$ =10 mm and $n$ =8. Scale bar: 50 mm.

Moreover, the origami spring crawlers have the ability to cross cliffs. Also using IOS-8 as the demonstration, the crawler climbed over a cliff step by step through properly stretching and contracting the body (Fig. 5c, Supplementary Movie S3). When the width of the cliff increased to a considerable value, the crawler stored elastic potential energy by stretching the body, and released it instantaneously to obtain enough initial velocity to leap over the cliff like an arrow (Fig. 5d, Supplementary Movie S3).

*3. Origami spring transformer.*

We demonstrate an origami spring transformer utilizing the deployability and the stretch-twist coupling. The fabrication process is exhibited in Fig. 6a, which follows three steps sequentially: connecting two IOS-4s ($r_0 = 10$ mm, $n = 4$) with opposite chirality as the "muscle", sealing the "muscle" by thin films as the skin, and mounting two wheels on both sides. The completed model is shown in Fig. 6b. By slowly supplying gas through the tube, the transformer can transform into three configurations owing to the extensibility and flexibility of IOSs, including a closed box, an upright body, and a two-wheeled car (Fig. 6c, Supplementary Movie S4). Furthermore, a rolling motion can be produced by employing the stretch-twist-coupling shape morphing of IOSs. As the shape morphing is symmetric with respect to the midplane, the rotation angle $\theta$ varying with the extension length $z$ can be calculated by Eq. (4) with $\theta = 4\tilde{\theta}$ (rad) and $z = 96.4\tilde{z}$ (mm). First, the transformer was in the closed box state, then the gas was rapidly supplied to stretch the "muscle", and the two wheels were thus pushed away. Meanwhile, the middle of the "muscle", namely the connection of two IOSs, underwent a twisting, leading the whole "car" to roll forward (Fig. 6d).

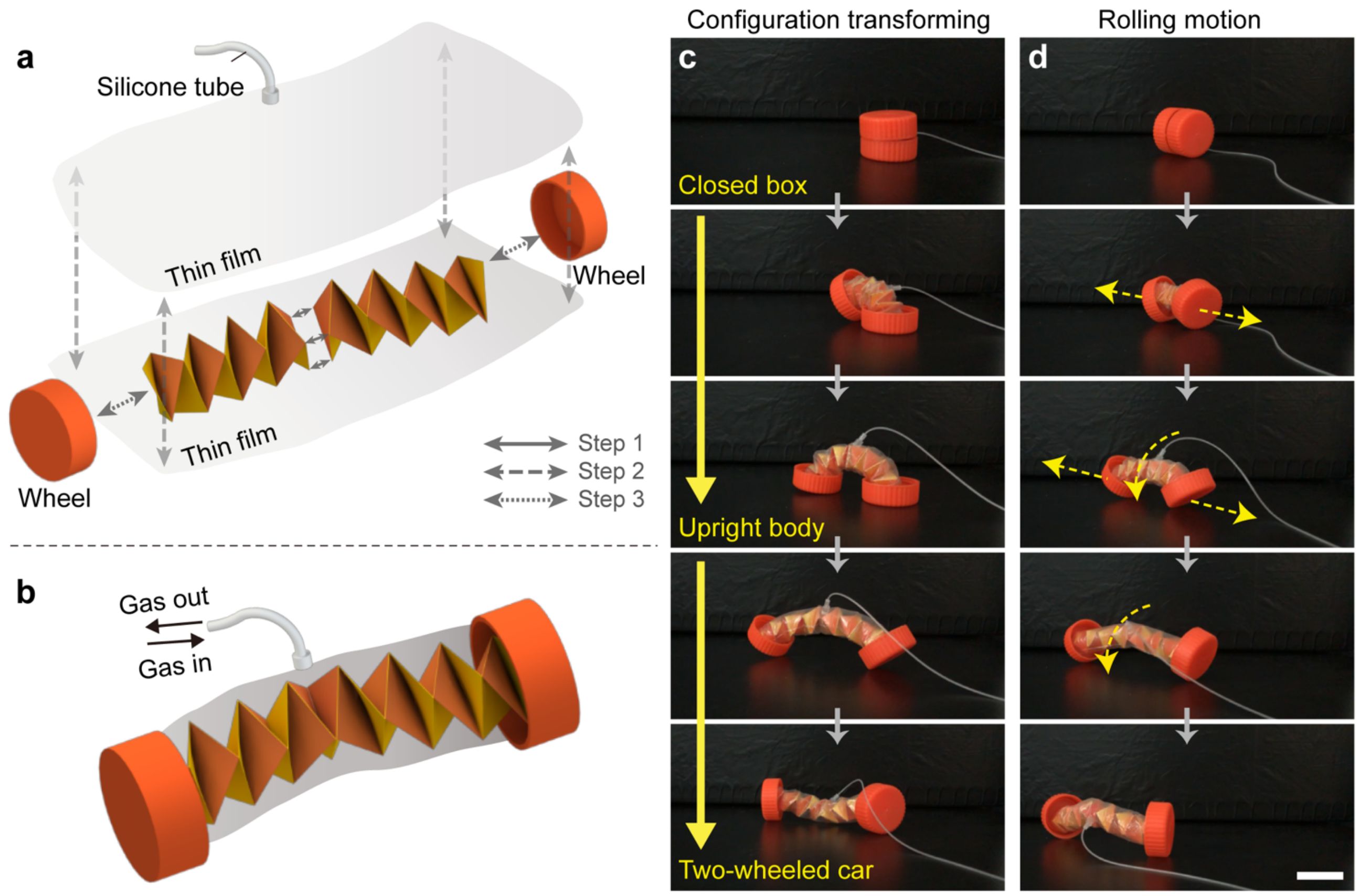

**FIG. 6.** Design, fabrication and resulting of an origami spring transformer. **(a)** The components introduction and fabrication process of the origami spring transformer. The process follows three steps, Step 1, 2, and 3 are indicated by solid-line, dashed-line, and dotted-line two-way arrows respectively. **(b)** The completed model. Gas is in or out via the tube to actuate the IOS. **(c)** The transforming process of origami spring transformer from a closed box to an upright body and then to a two-wheeled car by slowly stretching the "muscle". **(d)** The rolling motion of the origami spring transformer by rapidly stretching the "muscle", and utilizing the stretch-twist-coupling performance. Scale bar: 30 mm.

## Conclusion

In this work, we demonstrate that the structural design of generalized origami spring with good shape-morphing performance is of benefit to the applications of flexible robotics. We firstly present a design strategy of generalized interleaved origami springs (IOS), which exhibit stretch-twist coupling under uniaxial tension. We establish a continuum model and an origami rigidization method based on the experiments and finite element simulations to capture, analyze, and improve the shape-morphing performance of the IOSs. Guided by the theoretical framework, we propose ejectors and crawlers by employing the good elasticity of origami springs, and devise an origami spring transformer that is capable of multiple transformations and a rolling motion. Besides the IOS, the stretch-twist coupling has also been observed in some deployable cylindrical origami systems and found in wide applications.[22,36,37,41,42] However, comparing with them, IOS exhibits significant advantages in the simple and fast folding process, controllable buckling-folding transformation on facets, excellent stretch-twist-coupling characteristic, smooth and continuous folding/unfolding motions, and good maintaining of the central axis during shape morphing. These above-mentioned advantages indicate that IOSs are good candidates for structural designs of robotics.

The framework can be potentially and naturally generalized by using paper ribbons with different polygons or combining IOSs and RIOSs, so as to break the axially symmetric stretch-twist-coupling and achieve various shape-morphing performance. Alternatively, active materials can be used to design stimuli-sensitive origami springs. Taken together, we expect that our framework of origami-aided design will pave the way to facilitate the diverse shape morphing of flexible robotics.

## Author Disclosure Statement

No competing financial interests exist.

## Funding Information

This work was supported by the National Natural Science Foundation of China (NSFC) under grants nos. 91848201, 11988102, 11521202, 11872004, and 11802004 and Beijing Natural Science Foundation under grants no. L172002.

## Supplementary Material

Supplementary Note S1—S4

Supplementary Figure S1—S12

Supplementary Table S1—S3

Supplementary Information

# Origami spring-inspired shape morphing for flexible robotics

*Qianying Chen*[1,2], *Fan Feng*[1], *Pengyu Lv*[1], *Huiling Duan*[1,2*]

[1]State Key Laboratory for Turbulence and Complex Systems,
Department of Mechanics and Engineering Science, BIC-ESAT,
College of Engineering, Peking University, 100871 Beijing, China
[2]CAPT, HEDPS and IFSA Collaborative Innovation Center of MoE,
Peking University, 100871 Beijing, China
*Corresponding author. H.D.(Email:hlduan@pku.edu.cn)

## Note S1 Construction of generalized origami springs

Besides IOS-4, we also propose generalized origami springs by changing the square facets of IOS-4 into other regular-polygon shapes such as regular hexagon and regular octagon. IOS-6 is constructed by three paper ribbons consisting of a series of regular hexagons arrayed in chain. As shown in Figure S1, the three ribbons are placed in coincidence with the central axis 60 degrees apart, and then folded alternately in a similar way as IOS-4. IOS-8 is constructed by four paper ribbons consisting of a series of regular octagon arrayed in chain, while their central axes are 45 degrees apart, and the folded in the same way (Figure S2).

## Note S2 Derivations of geometrical parameters

### Note S2.1 Radius of IOS-4

Assume the edges AB, BC, CD and DA in Figure S3 are all straight lines with lengths $a$. The $z$-axis coordinates of the vertices on the curved surface ABCD shown in Figure S3 can be expressed as

$$\begin{cases} z_{\mathrm{A}} = z_{\mathrm{A}}, \\ z_{\mathrm{B}} = z_{\mathrm{A}} + a\sin\varphi, \\ z_{\mathrm{C}} = z_{\mathrm{A}} + \dfrac{2a}{3}\sin\varphi, \\ z_{\mathrm{D}} = z_{\mathrm{A}} + \dfrac{a}{3}\sin\varphi = z_{\mathrm{A}} + a\sin\varphi'. \end{cases} \tag{S1}$$

where $\varphi'$ and $\varphi$ are the helix angles of the small-pitch helix and the large-pitch helix respectively. Therefore, one can obtain the relationship between $\varphi$ and $\varphi'$ as

$$\varphi' = \arcsin\left(\frac{1}{3}\sin\varphi\right) = \arcsin\frac{\tilde{z}}{3}, \quad \varphi \in [0, \frac{\pi}{2}), \quad \varphi' \in [0, \arcsin\frac{1}{3}). \tag{S2}$$

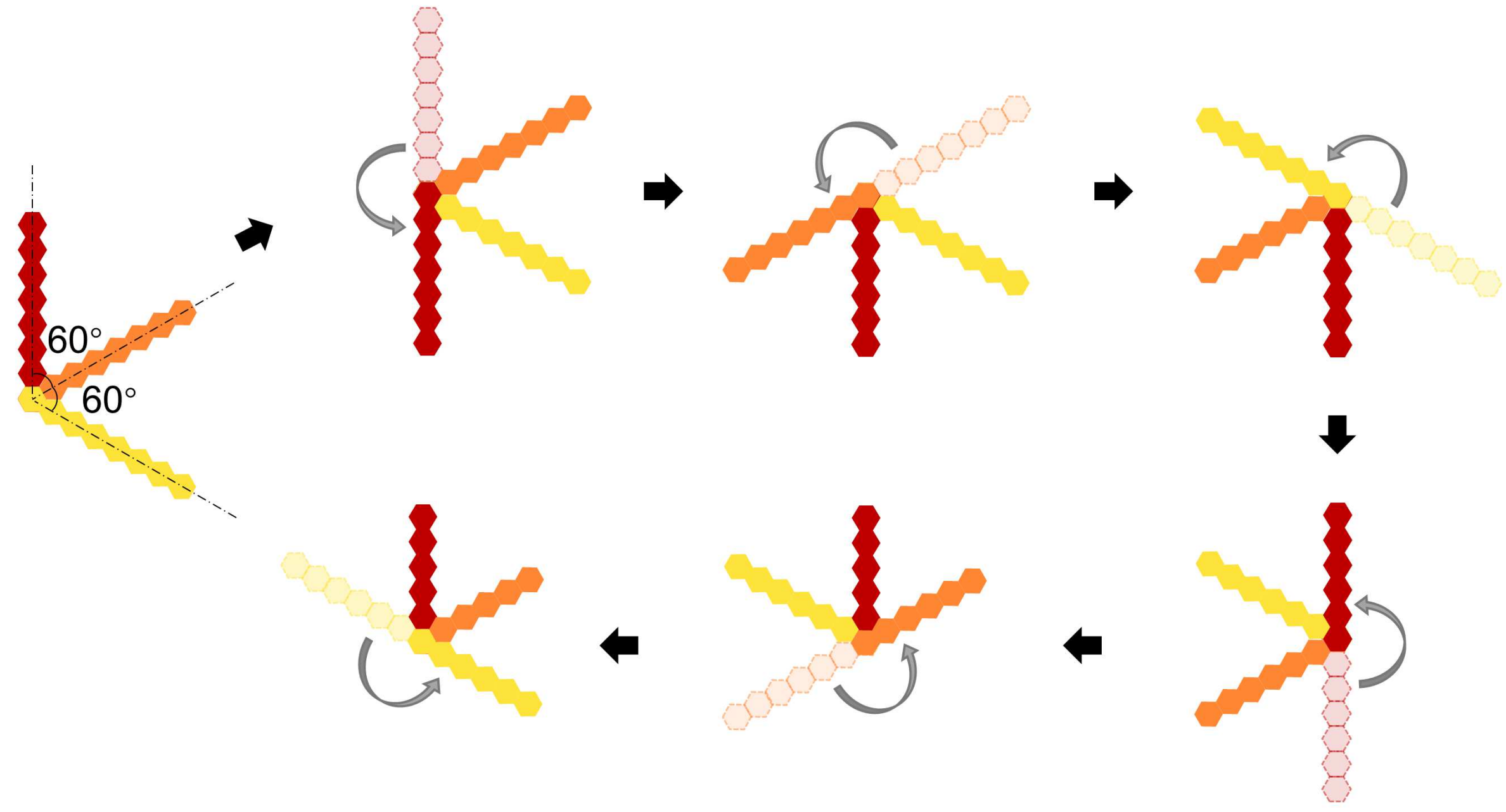

Figure S1: Folding process of IOS-6.

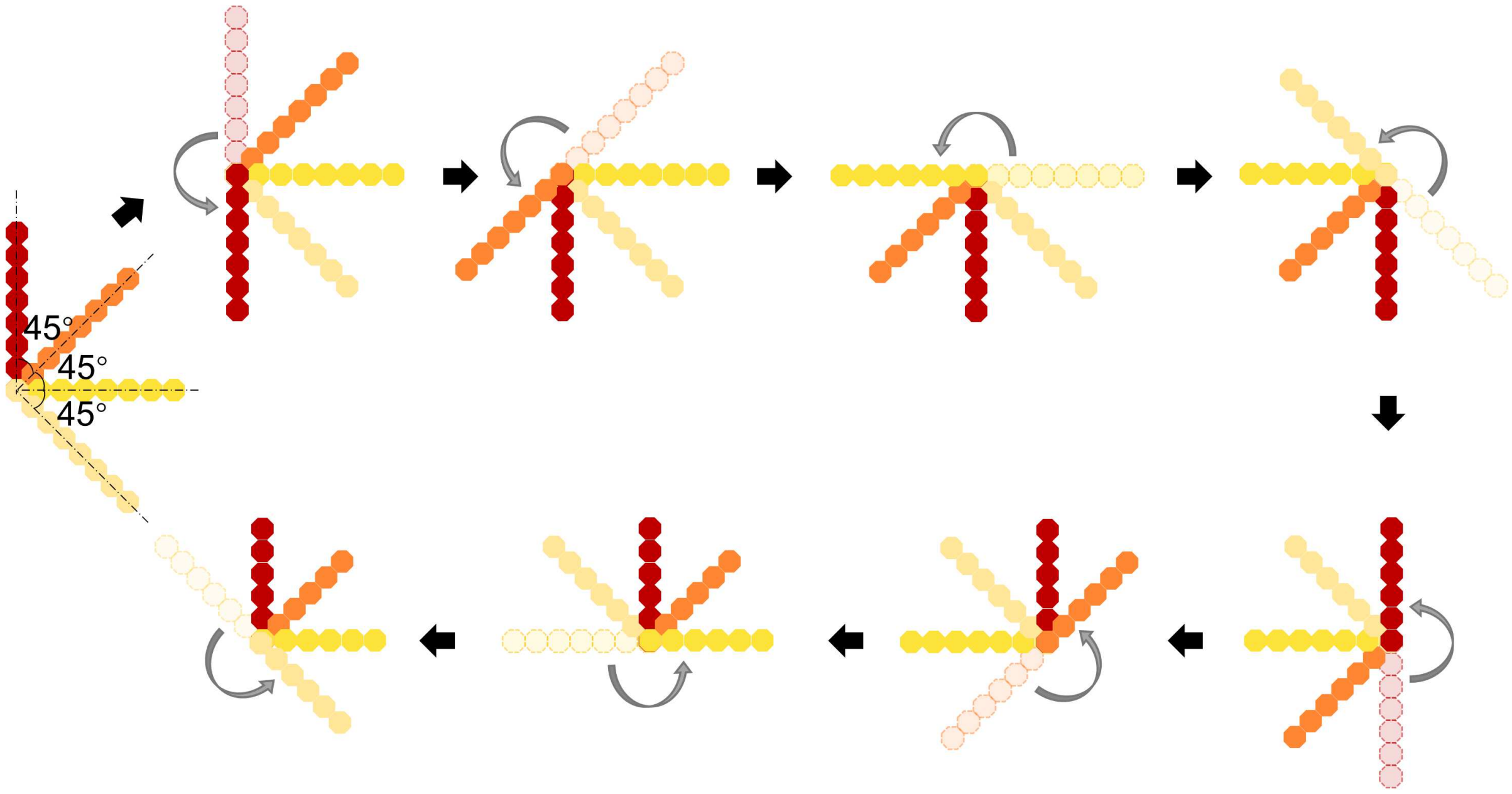

Figure S2: Folding process of IOS-8.

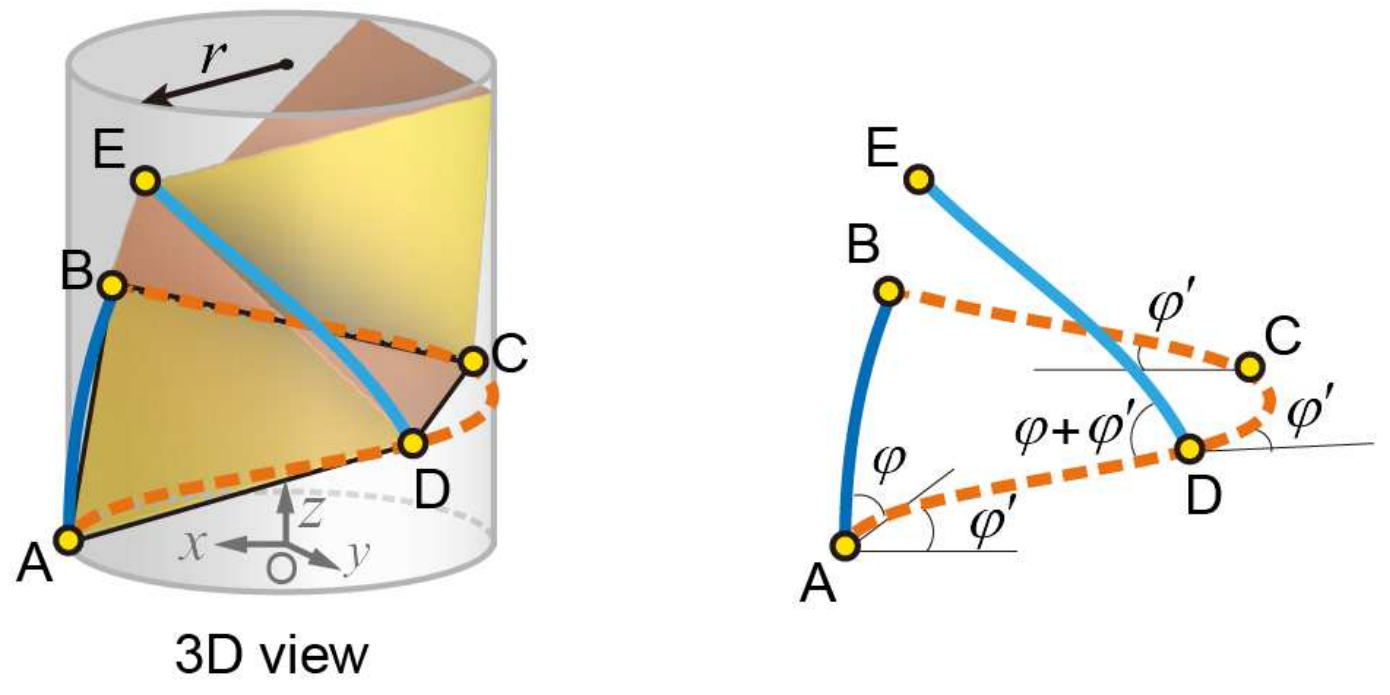

Figure S3: Schematic of the IOS-4 unit cell. A, D, C and B are sequentially placed on the small-pitch helix with helix angle $\varphi'$. A and B are on the large-pitch helix with helix angle $\varphi$. The folding angle along the primary crease is approximated by $\varphi+\varphi'$, i.e., $\angle$ EDA. Similar for IOS-6 and IOS-8.

Further, the arc lengths of each segment on the projected circle AB′C′D′ (Fig. 2b in the main text) and their relations are deduced unitizing geometric analysis

$$\widehat{\mathrm{AB'}} = r\arccos\frac{2r^2-a^2\cos^2\varphi}{2r^2}, \tag{S3}$$

$$\widehat{\mathrm{B'C'}} = \widehat{\mathrm{C'D'}} = \widehat{\mathrm{D'A'}} = r\arccos\frac{2r^2-a^2\cos^2\varphi'}{2r^2}, \tag{S4}$$

$$r\arccos\frac{2r^2-a^2\cos^2\varphi}{2r^2} + 3r\arccos\frac{2r^2-a^2\cos^2\varphi'}{2r^2} = 2\pi r. \tag{S5}$$

Therefore, the expression of the radius $r$ can be deduced as

$$\frac{r}{r_0} = \frac{N_4}{6}\sqrt{\frac{N_4(M_4+N_4)}{3}}, \tag{S6}$$

where $M_4=\sqrt{1-\tilde{z}^2}$, $N_4=\sqrt{9-\tilde{z}^2}$, $\tilde{z}=\sin\varphi$, and $r_0=\sqrt{2}a/2$ is the circumcircle radius of the square with side length $a$.

### Note S2.2 Radius of IOS-6

Similar to the derivation process of IOS-4, the $z$-axis coordinates of the vertices on the curved surface ABCDEF shown in Figure S4 can be expressed as

$$\begin{cases} z_\mathrm{A} = z_\mathrm{A}, \\ z_\mathrm{B} = z_\mathrm{A} + a\sin\varphi, \\ z_\mathrm{C} = z_\mathrm{A} + 2a\sin\varphi, \\ z_\mathrm{D} = z_\mathrm{A} + \dfrac{3a}{2}\sin\varphi, \\ z_\mathrm{E} = z_\mathrm{A} + a\sin\varphi, \\ z_\mathrm{F} = z_\mathrm{A} + \dfrac{a}{2}\sin\varphi = z_\mathrm{A} + a\sin\varphi'. \end{cases} \tag{S7}$$

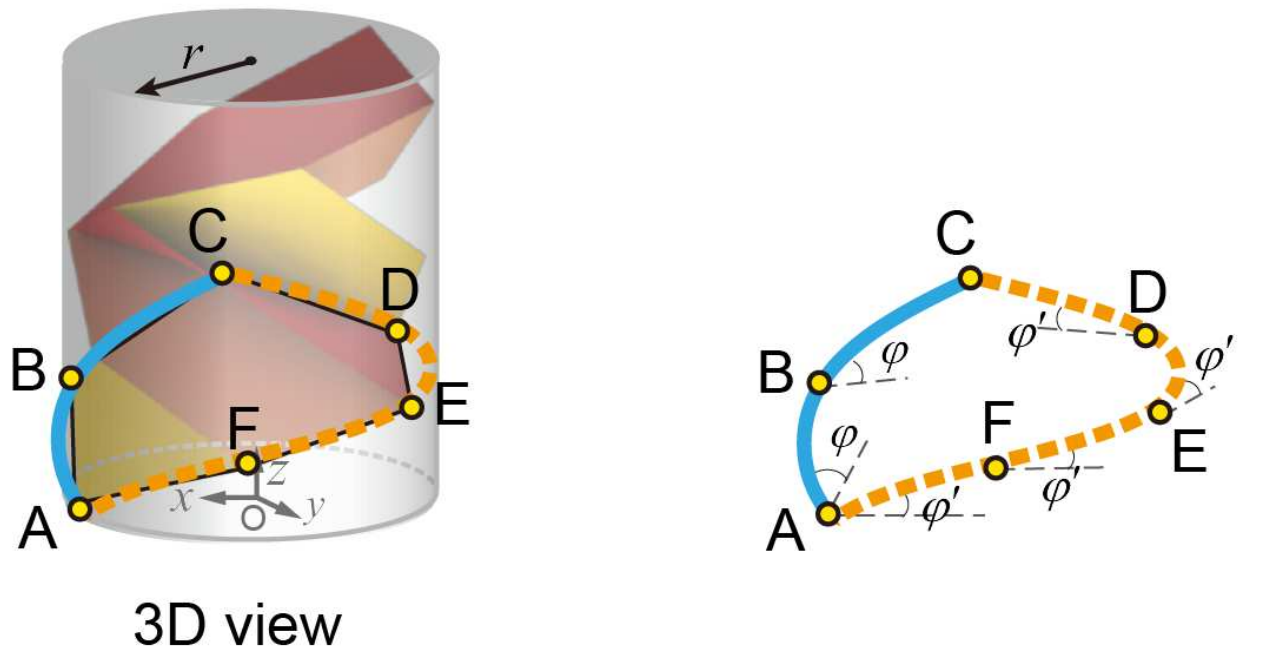

Figure S4: Schematic of the IOS-6 unit cell. A, F, E, D and C are sequentially placed on the small-pitch helix with helix angle $\varphi'$. A, B and C are on one of the large-pitch helices with helix angle $\varphi$.

Therefore we have

$$\varphi' = \arcsin\left(\frac{1}{2}\sin\varphi\right) = \arcsin\frac{\tilde{z}}{2}, \quad \varphi \in [0, \frac{\pi}{2}), \quad \varphi' \in [0, \arcsin\frac{1}{2}). \tag{S8}$$

Further, employing the similar calculation in IOS-4, the circumference of the cylinder is expressed as

$$2\pi r = 2r\arccos\frac{2r^2 - a^2\cos^2\varphi}{2r^2} + 4r\arccos\frac{2r^2 - a^2\cos^2\varphi'}{2r^2}, \tag{S9}$$

and thus

$$\frac{r}{r_0} = \frac{1}{2}\sqrt{\frac{N_6(M_6 + 2N_6) - 1}{2}}, \tag{S10}$$

where $M_6 = \sqrt{3 - 3\tilde{z}^2}$, $N_6 = \sqrt{3 - \tilde{z}^2}$, and $r_0 = a$ is the circumcircle radius of the hexagon with side length $a$.

### Note S2.3 Radius of IOS-8

The $z$-axis coordinates of the vertices on the curved surface ABCDEFGH shown in Figure S5 can be expressed as

$$
\begin{cases}
z_A = z_A, \\
z_B = z_A + a\sin\varphi, \\
z_C = z_A + 2a\sin\varphi, \\
z_D = z_A + 3a\sin\varphi = z_A + 5a\sin\varphi', \\
z_E = z_A + 4a\sin\varphi', \\
z_F = z_A + 3a\sin\varphi', \\
z_G = z_A + 2a\sin\varphi', \\
z_H = z_A + a\sin\varphi'.
\end{cases} \tag{S11}
$$

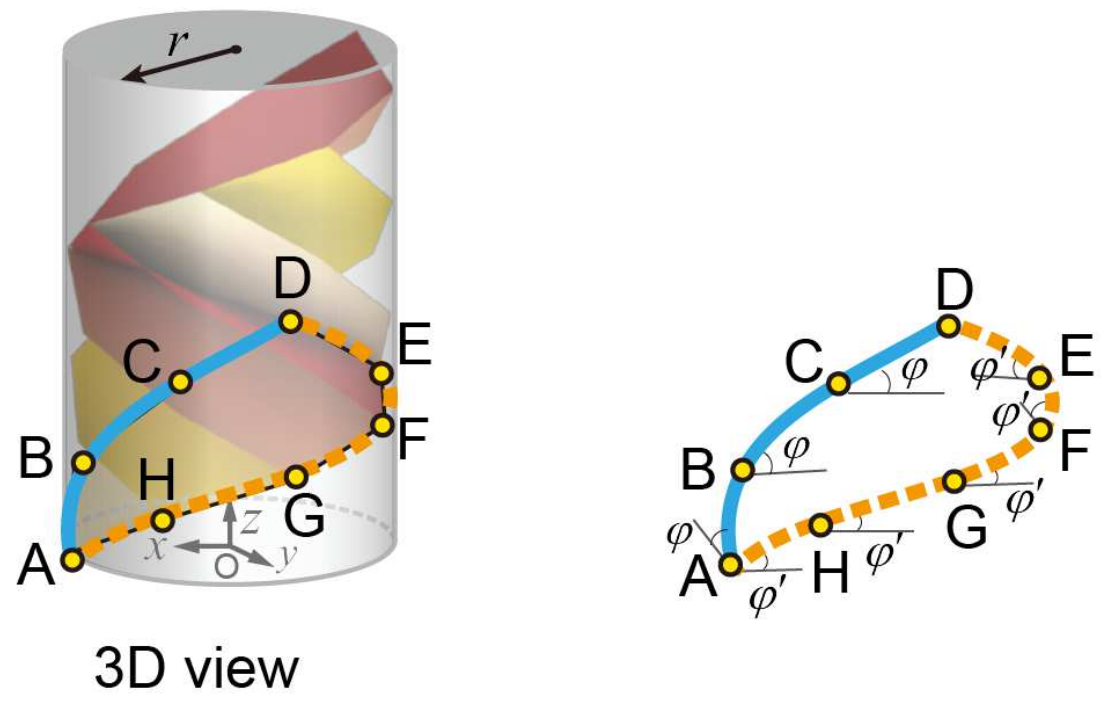

Figure S5: Schematic of the IOS-8 unit cell. A, H, G, F, E and D are sequentially placed on the small-pitch helix with helix angle $\varphi'$. A, B, C and D are on one of the large-pitch helices with helix angle $\varphi$.

Therefore, we have

$$
\varphi' = \arcsin\frac{3}{5}\sin\varphi = \arcsin\frac{3\tilde{z}}{5}, \quad \varphi \in [0, \frac{\pi}{2}), \quad \varphi' \in [0, \arcsin\frac{3}{5}). \tag{S12}
$$

Further, by the similar calculation in IOS-4, the circumference of the cylinder is

$$
2\pi r = 3r\arccos\frac{2r^2 - a^2\cos^2\varphi}{2r^2} + 5r\arccos\frac{2r^2 - a^2\cos^2\varphi'}{2r^2}, \tag{S13}
$$

and thus

$$
\frac{r}{r_0} = \frac{N_8}{5}\sqrt{\frac{2(2-\sqrt{2})}{s_{MN}}}, \tag{S14}
$$

where $M_8 = 5\sqrt{1-\tilde{z}^2}$, $N_8 = \sqrt{25 - 9\tilde{z}^2}$, $s_{MN} = 5 - l_{MN}^3 - (1 + l_{MN} - l_{MN}^2)\sqrt{5 + 2l_{MN} + l_{MN}^2}$, $l_{MN} = M_8/N_8$, and $r_0 = \sqrt{1 + 1/\sqrt{2}}a$ is the circumcircle of the octagon with side length $a$.

The curves of $r/r_0$ varying with extension ratio $\tilde{z}$ of IOS-4, IOS-6 and IOS-8 are shown in Figure S6, which all exhibit a descreasing trend with the increasing $\tilde{z}$, indicating that the radius of the cylinder decreases as the extension progresses.

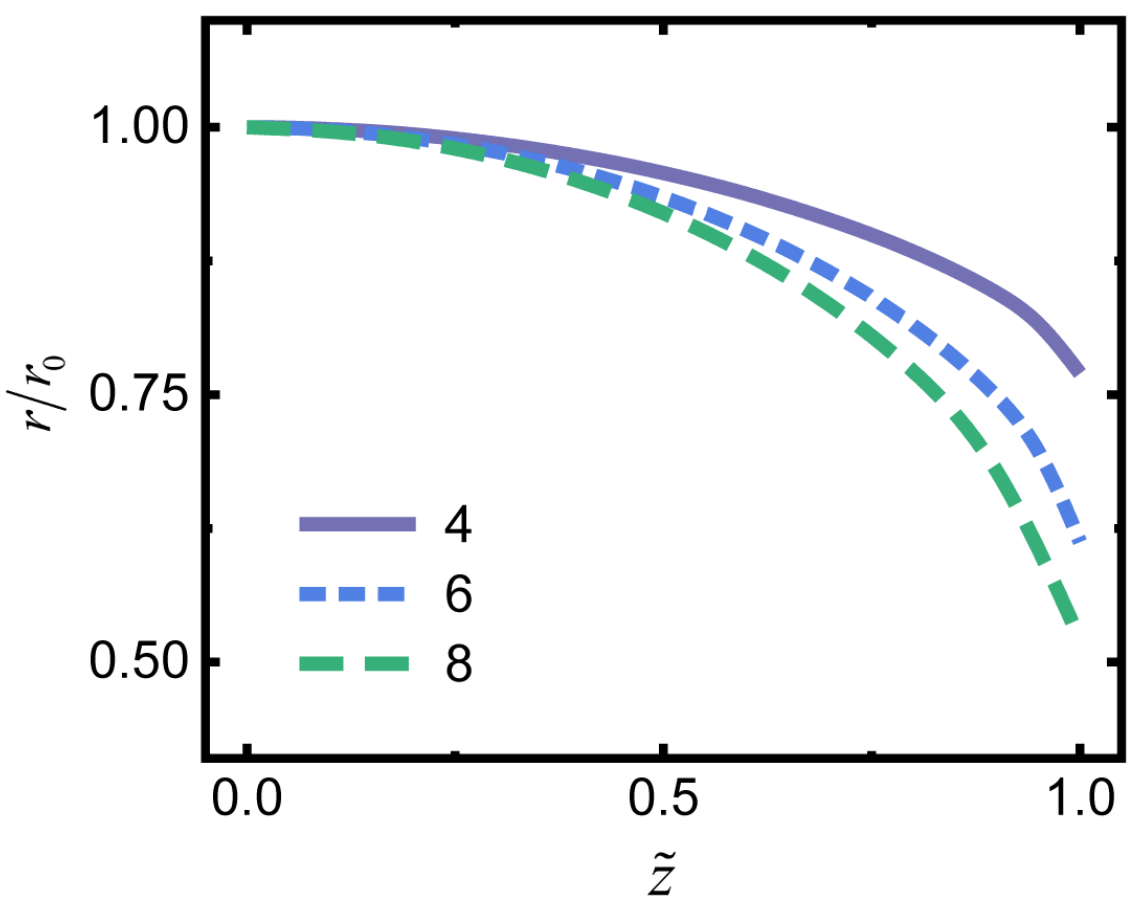

Figure S6: $r/r_0$ versus the extension ratio $\tilde{z}$.

## Note S2.4 Folding angles of IOS-4 and RIOS-4

The coordinates $(x_i, y_i, z_i)$ of the vertices on the curved surface *ABCD* shown in Fig. 2b can be expressed as

$$\begin{aligned} &\mathrm{A}(r,0,0), \quad \mathrm{B}(r\cos\beta_\mathrm{B}, r\sin\beta_\mathrm{B}, a\tilde{z}), \\ &\mathrm{C}(r\cos\beta_\mathrm{C}, r\sin\beta_\mathrm{C}, \frac{2}{3}a\tilde{z}), \quad \mathrm{D}(r\cos\beta_\mathrm{D}, r\sin\beta_\mathrm{D}, \frac{1}{3}a\tilde{z}), \end{aligned} \tag{S15}$$

where $\beta_\mathrm{B} = 2\pi - P_4$, $\beta_\mathrm{C} = 2Q_4$, $\beta_\mathrm{D} = Q_4$. See $P_4$ and $Q_4$ in Table S1 for details. The $x$ and $y$ coordinates are obtained by projecting the vertices onto the 2D plane perpendicular to the cylindrical axis, and the $z$ coordinates are calculated from the helix angle and the side length. We summarize these geometrical quantities in Table S1 with no details of calculation for simplicity. Similar for IOS-6 and IOS-8. All the coordinates are functions of $\tilde{z}$, and therefore the geometrical quantities deduced from these coordinates are also unitary functions of $\tilde{z}$.

We simplify the buckling of IOS-4 to two equal folding angles $\omega_\mathrm{B}$ along the diagonal. As shown in Figure S7a, $\omega_\mathrm{B}$ is calculated by

$$\omega_\mathrm{B} = \pi - \angle \mathrm{AO_{ABCD}C} = \pi - \angle \mathrm{BO_{ABCD}D}, \tag{S16}$$

where $O_\mathrm{ABCD}((x_\mathrm{A}+x_\mathrm{B}+x_\mathrm{C}+x_\mathrm{D})/4, (y_\mathrm{A}+y_\mathrm{B}+y_\mathrm{C}+y_\mathrm{D})/4, (z_\mathrm{A}+z_\mathrm{B}+z_\mathrm{C}+z_\mathrm{D})/4)$ is the center of the surface ABCD. $\omega_\mathrm{B}$ is a unary function of $\tilde{z}$ and can be deduced by coordinate evaluating.

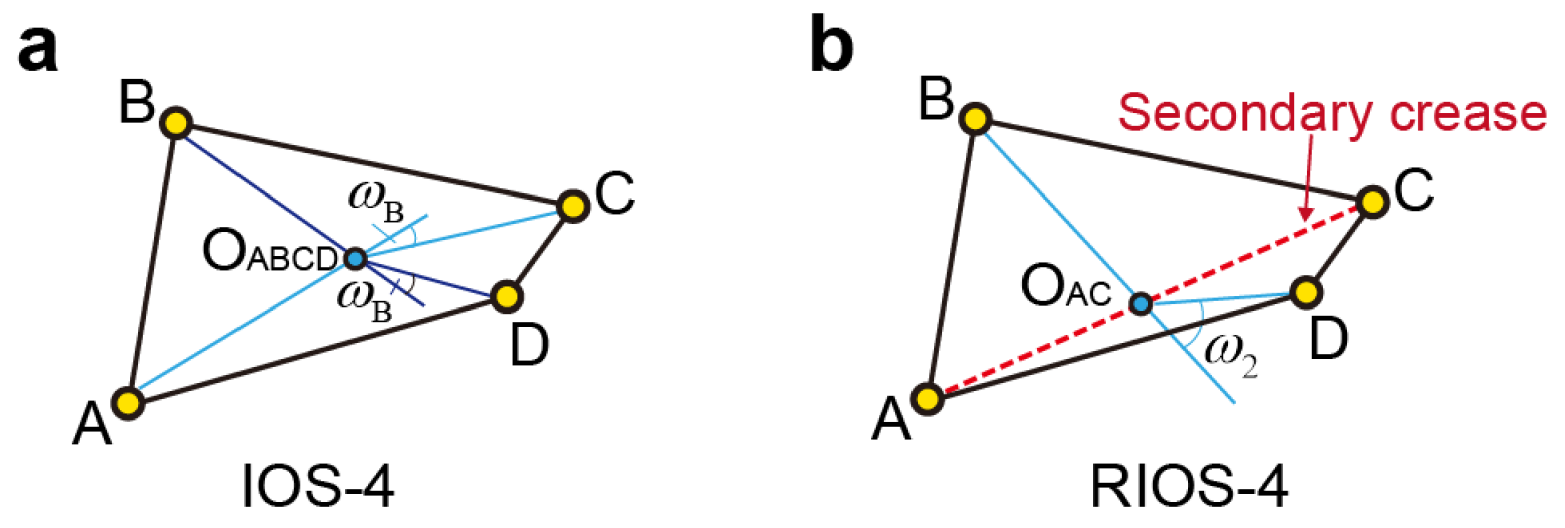

Figure S7: Diagram of the folding angles of IOS-4 and RIOS-4. (a) $\omega_\mathrm{B}$ of IOS-4. (b) $\omega_2$ of RIOS-4.

The folding angle $\omega_2$ along the secondary crease of RIOS-4 (Figure S7b) can be expressed as

$$\omega_2 = \pi - \angle \mathrm{BO_{AC}D}, \tag{S17}$$

where $O_\mathrm{AC}((x_\mathrm{A}+x_\mathrm{C})/2, (y_\mathrm{A}+y_\mathrm{C})/2, (z_\mathrm{A}+z_\mathrm{C})/2)$ is the middle point of AC. $\omega_2$ is a unary function of $\tilde{z}$.

## Note S2.5 Folding angles of IOS-6 and RIOS-6

We have the coordinates of the vertices as

$$\begin{aligned} &\mathrm{A}(r,0,0), \quad \mathrm{B}(r\cos\beta_\mathrm{B}, r\sin\beta_\mathrm{B}, a\tilde{z}), \\ &\mathrm{C}(r\cos\beta_\mathrm{C}, r\sin\beta_\mathrm{C}, 2a\tilde{z}), \quad \mathrm{D}(r\cos\beta_\mathrm{D}, r\sin\beta_\mathrm{D}, \frac{3}{2}a\tilde{z}), \\ &\mathrm{E}(r\cos\beta_\mathrm{E}, r\sin\beta_\mathrm{E}, a\tilde{z}), \quad \mathrm{F}(r\cos\beta_\mathrm{F}, r\sin\beta_\mathrm{F}, \frac{1}{2}a\tilde{z}), \end{aligned} \tag{S18}$$

where $\beta_B = 2\pi - P_6$, $\beta_C = 4Q_6$, $\beta_D = 3Q_6$, $\beta_E = 2Q_6$, $\beta_F = Q_6$. See $P_6$ and $Q_6$ in Table S1 for detail. Therefore, the folding angle $\omega_B$ (Figure S8a) is deduced as

$$\omega_B = \pi - \angle AO_{ACDF}D = \pi - \angle CO_{ACDF}F, \tag{S19}$$

where $O_{ACDF}((x_A + x_C + x_D + x_F)/4, (y_A + y_C + y_D + y_F)/4, (z_A + z_C + z_D + z_F)/4)$ is the center point of ACDF. Similarly, $\omega_B$ is a unary function of $\tilde{z}$.

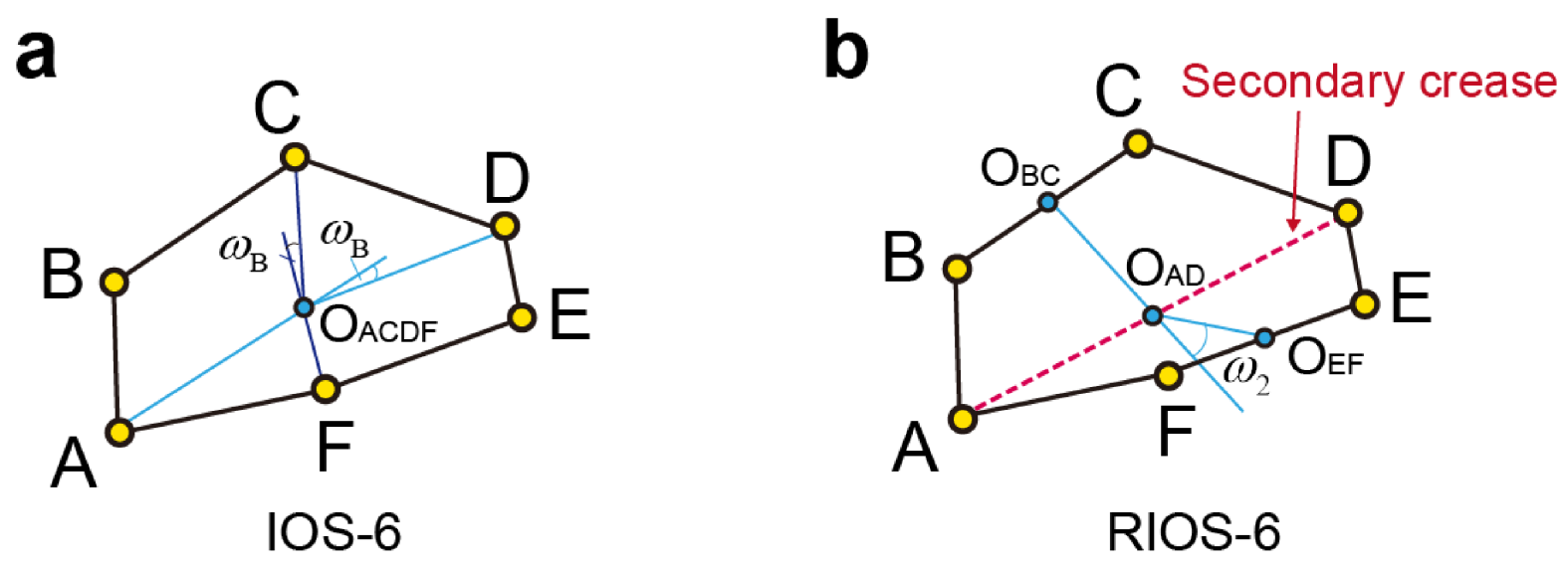

Figure S8: Diagram of the folding angles of IOS-6 and RIOS-6. (a) $\omega_B$ of IOS-6. (b) $\omega_2$ of RIOS-6.

The folding angle $\omega_2$ along the secondary crease of RIOS-6 (Figure S8b) can be expressed as

$$\omega_2 = \pi - \angle O_{BC}O_{AD}O_{EF}, \tag{S20}$$

where $O_{BC}((x_B + x_C)/2, (y_B + y_C)/2, (z_B + z_C)/2)$, $O_{AD}((x_A + x_D)/2, (y_A + y_D)/2, (z_A + z_D)/2)$, $O_{EF}((x_E + x_F)/2, (y_E + y_F)/2, (z_E + z_F)/2)$ are the middle points of CB, AD and EF respectively. Again, $\omega_2$ is a unary function of $\tilde{z}$.

## Note S2.6 Folding angles of IOS-8 and RIOS-8

We have the coordinates of the vertices as

$$\begin{aligned} &A(r,0,0), \quad B(r\cos\beta_B, r\sin\beta_B, a\tilde{z}), \\ &C(r\cos\beta_C, r\sin\beta_C, 2a\tilde{z}), \quad D(r\cos\beta_D, r\sin\beta_D, 3a\tilde{z}), \\ &E(r\cos\beta_E, r\sin\beta_E, \frac{12}{5}a\tilde{z}), \quad F(r\cos\beta_F, r\sin\beta_F, \frac{9}{5}a\tilde{z}), \\ &G(r\cos\beta_G, r\sin\beta_G, \frac{6}{5}a\tilde{z}), \quad H(r\cos\beta_H, r\sin\beta_H, \frac{3}{5}a\tilde{z}), \end{aligned} \tag{S21}$$

where $\beta_B = 2\pi - P_8$, $\beta_C = 2\pi - 2P_8$, $\beta_D = 5Q_8$, $\beta_E = 4Q_8$, $\beta_F = 3Q_8$, $\beta_G = 2Q_8$, $\beta_H = Q_8$. See $P_8$ and $Q_8$ in Table S1 for detail. Therefore, the folding angle $\omega_B$ (Figure S9a) is deduced as

$$\omega_B = \pi - \angle AO_{ADEH}E = \pi - \angle DO_{ADEH}H, \tag{S22}$$

where $O_{ADEH}((x_A + x_D + x_E + x_H)/4, (y_A + y_D + y_E + y_H)/4, (z_A + z_D + z_E + z_H)/4)$ is the center point of ADEH. $\omega_B$ is a unary function of $\tilde{z}$.

The folding angle $\omega_2$ along the secondary crease of RIOS-8 (Figure S9b) can be expressed as

$$\omega_2 = \pi - \angle CO_{AE}G, \tag{S23}$$

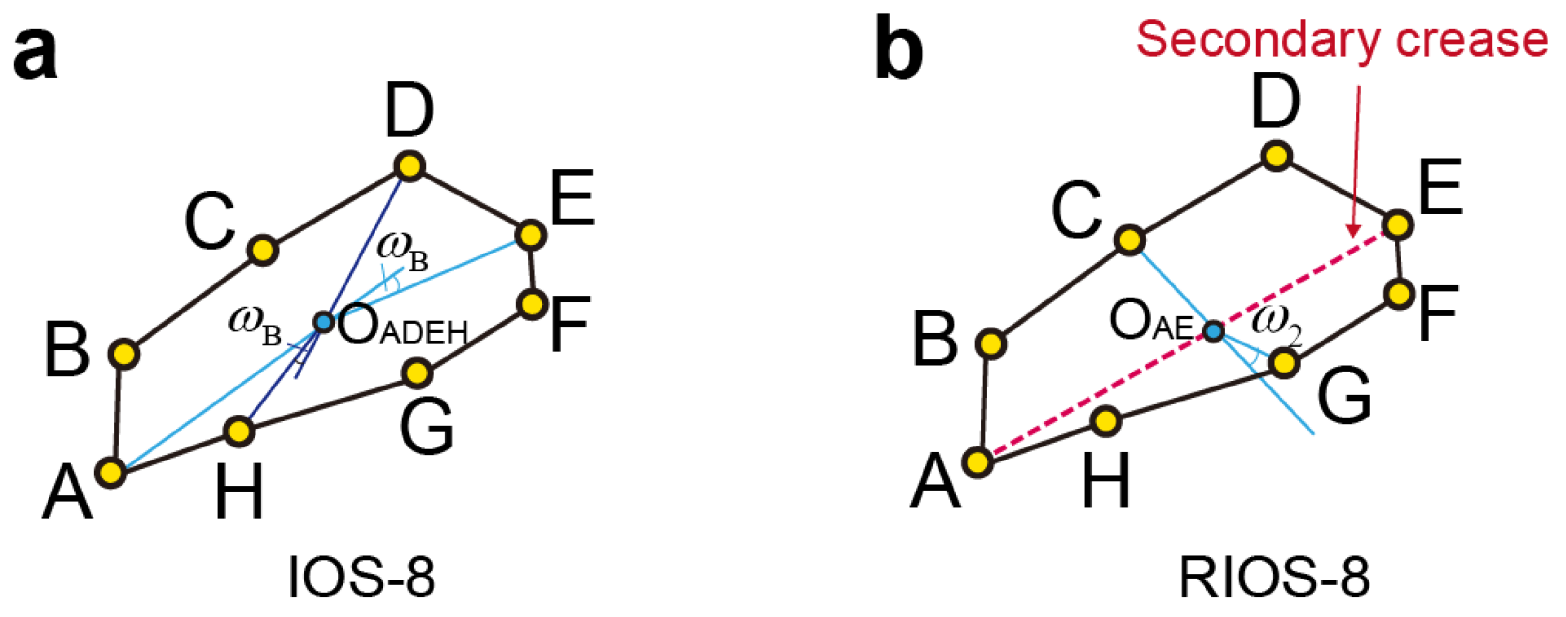

Figure S9: Diagram of the folding angles of IOS-8 and RIOS-8. (a) $\omega_B$ of IOS-8. (b) $\omega_2$ of RIOS-8.

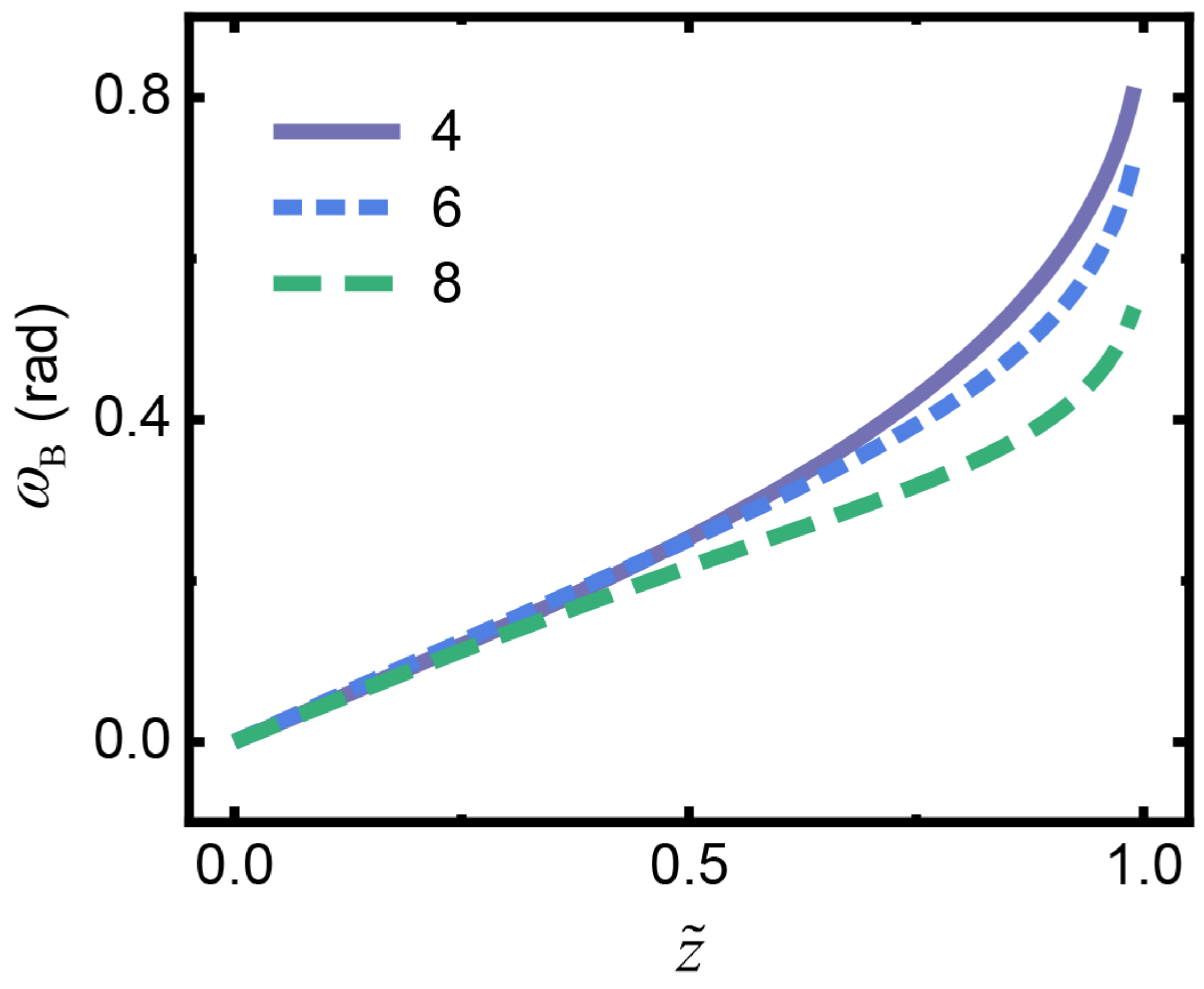

Figure S10: $\omega_B$ versus the extension ratio $\tilde{z}$.

where $O_{AE}((x_A + x_E)/2, (y_A + y_E)/2, (z_A + z_E)/2)$ is the middle point of AE. $\omega_2$ is a unary function of $\tilde{z}$.

The curves of $\omega_B$ and $\omega_2$ varying with extension ratio $\tilde{z}$ of different types of origami springs are shown in Figure S10 and Figure S11, respetively, which all exhibit an increasing trend with the increasing $\tilde{z}$.

## Note S3 Mechanical testing

We tested the mechanical properties of the paper for origami springs and the structures of the proposed origami springs by MTESTQuatro Testing System (ADMET, Norwood, MA). The paper for origami was cut into a dog-bone-shape standard specimen and underwent a uniaxial tension with 0.01-mm/s loading rate. We regarded the mechanical response of the paper as linear elasticity and the Young's modulus was obtain about 1738.3 MPa after multiple measurements. IOSs and RIOSs were pulled with fine threads drawing from the midpoints of both end facets. Uniaxial tension and compression were applied both from the free extension state of each origami spring with 0.01-mm/s loading rate. The force-extension curves are exhibited in Fig. 3c in the main text.

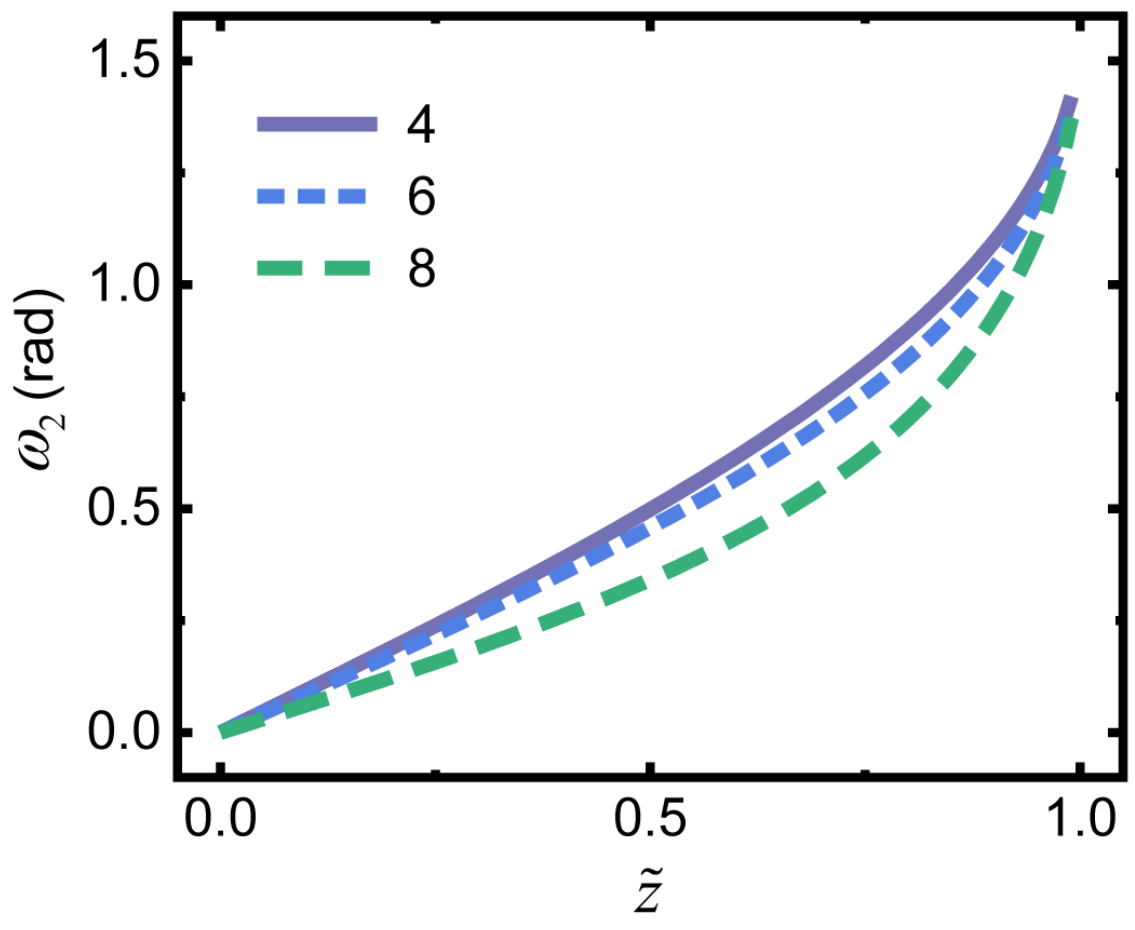

Figure S11: $\omega_2$ versus the extension ratio $\tilde{z}$.

## Note S4 Finite element analysis (FEA)

We carried out the FEA using the commercial finite element software ABAQUS. We first set up a 3D deformable solid part of the folding zigzag origami in square (or regular hexagon, or regular octagon) facets, eight unit cells, 10-mm circumcircle radius, and 0.1-mm paper thickness. The Young's modulus and the Poisson's ratio were set to 1738.3 MPa and 0.3, respectively. Then the zigzag origami structures were assembled with proper relative positions to construct origami springs. The boundaries of facets and the creases in contact were tied together to constraint their relative displacements. The center region of the bottom facet was pinned and the center region of the top facet was applied a 150-mm displacement constraint. 3D models of the origami springs were meshed using the element type C3D8R (8-node linear brick, reduced integration), while proper mesh sizes were selected to ensure the computational convergence and accuracy.

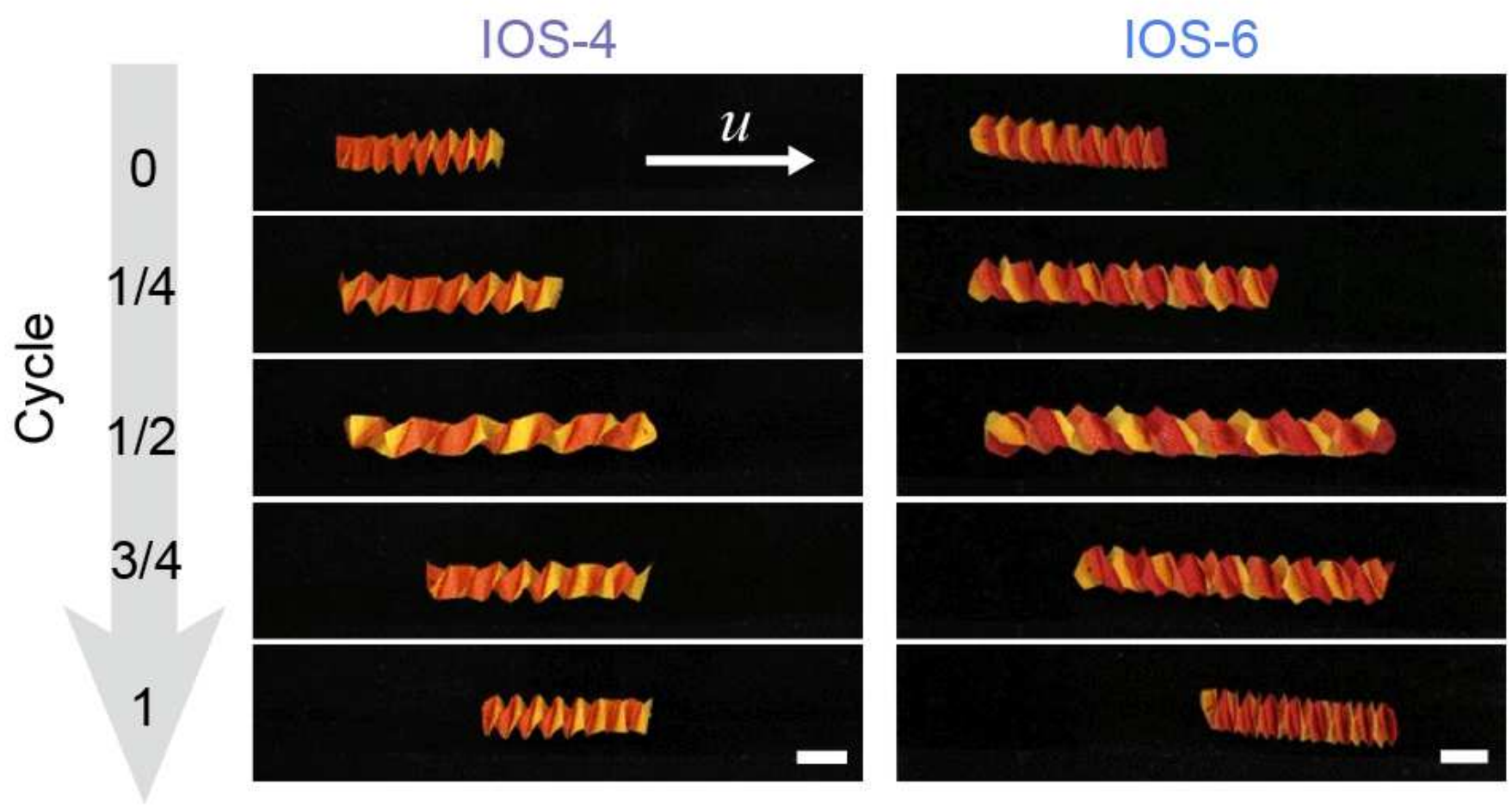

Figure S12: Five typical crawling moments of IOS-4 and IOS-6 throughout an entire crawling cycle. Scale bar: 20 mm.

Table S1: Important geometric parameters of IOSs in different types.

| **Types** | **IOS-4** | **IOS-6** | **IOS-8** |
|---|---|---|---|
| $a$ | $\sqrt{2}r_0$ | $r_0$ | $\sqrt{2-\sqrt{2}}r_0$ |
| $\varphi$ | $\arcsin\tilde{z}$ | $\arcsin\tilde{z}$ | $\arcsin\tilde{z}$ |
| $\varphi'$ | $\arcsin\frac{\tilde{z}}{3}$ | $\arcsin\frac{\tilde{z}}{2}$ | $\arcsin\frac{3\tilde{z}}{5}$ |
| $\frac{r}{r_0}$ | $\frac{N_4}{6}\sqrt{\frac{N_4(M_4+N_4)}{3}}$ | $\frac{1}{2}\sqrt{\frac{N_6(M_6+2N_6)-1}{2}}$ | $\frac{N_8}{5}\sqrt{\frac{2(2-\sqrt{2})}{s_{\mathrm{MN}}}}$ |
| $\tilde{\theta}$ | $4\arccos\left(P_4Q_4+\sqrt{1-P_4^2}\sqrt{1-Q_4^2}\right)$ | $6\arccos\left(P_6Q_6+\sqrt{1-P_6^2}\sqrt{1-Q_6^2}\right)$ | $8\arccos\left(P_8Q_8+\sqrt{1-P_8^2}\sqrt{1-Q_8^2}\right)$ |
| $l_{\mathrm{f}}$ | $2an$ | $2\sqrt{3}an$ | $2(\sqrt{2}+1)an$ |
| $\lambda$ | 1.205 | 1.110 | 1.007 |
| | $M_4=\sqrt{1-\tilde{z}^2}$, $N_4=\sqrt{9-\tilde{z}^2}$, $P_4=1-\frac{108M_4^2}{N_4^3(M_4+N_4)}$, $Q_4=1-\frac{12}{N_4(M_4+N_4)}$. | $M_6=\sqrt{3-3\tilde{z}^2}$, $N_6=\sqrt{3-\tilde{z}^2}$, $P_6=1-\frac{4M_6^2}{3N_6(M_6+2N_6)-3}$, $Q_6=1-\frac{N_6^2+1}{N_6(M_6+2N_6)-1}$. | $M_8=5\sqrt{1-\tilde{z}^2}$, $N_8=\sqrt{25-9\tilde{z}^2}$, $P_8=1-\frac{l_{\mathrm{MN}}^2 s_{\mathrm{MN}}}{4}$, $Q_8=1-\frac{s_{\mathrm{MN}}}{4}$, $l_{\mathrm{MN}}=M_8/N_8$, $s_{\mathrm{MN}}=5-l_{\mathrm{MN}}^3-(1+l_{\mathrm{MN}}-l_{\mathrm{MN}}^2)\sqrt{5+2l_{\mathrm{MN}}+l_{\mathrm{MN}}^2}$. |

Table S2: Mechanical parameters of IOSs and RIOSs in different types.

| Types | IOS-4 | RIOS-4 | IOS-6 | RIOS-6 | IOS-8 | RIOS-8 |
|---|---|---|---|---|---|---|
| $A_1$ | $(\varphi+\varphi')\frac{\partial(\varphi+\varphi')}{\partial\tilde{z}}$ | | | | | |
| $A_2$ | $\frac{(\varphi+\varphi')^2}{2\tilde{z}}$ | | | | | |
| $B_1$ | $2\sqrt{2}\omega_{\mathrm{B}}\frac{\partial\omega_{\mathrm{B}}}{\partial\tilde{z}}$ | - | $4\omega_{\mathrm{B}}\frac{\partial\omega_{\mathrm{B}}}{\partial\tilde{z}}$ | - | $(2^{\frac{7}{4}}+2^{\frac{9}{4}})\omega_{\mathrm{B}}\frac{\partial\omega_{\mathrm{B}}}{\partial\tilde{z}}$ | - |
| $B_2$ | $\frac{\sqrt{2}\omega_{\mathrm{B}}^2}{\tilde{z}}$ | - | $\frac{2\omega_{\mathrm{B}}^2}{\tilde{z}}$ | - | $(2^{\frac{3}{4}}+2^{\frac{5}{4}})\frac{\omega_{\mathrm{B}}^2}{\tilde{z}}$ | - |
| $C_1$ | - | $\sqrt{2}\omega_2\frac{\partial\omega_2}{\partial\tilde{z}}$ | - | $2\omega_{\mathrm{B}}\frac{\partial\omega_{\mathrm{B}}}{\partial\tilde{z}}$ | - | $(2^{\frac{3}{4}}+2^{\frac{4}{4}})\omega_2\frac{\partial\omega_2}{\partial\tilde{z}}$ |
| $C_2$ | - | $\frac{\sqrt{2}\omega_2^2}{2\tilde{z}}$ | - | $\frac{\omega_2^2}{\tilde{z}}$ | - | $(2^{-\frac{1}{4}}+2^{\frac{1}{4}})\frac{\omega_{\mathrm{B}}^2}{\tilde{z}}$ |
| $\tilde{z}_0$ | 0.242 | 0.385 | 0.241 | 0.319 | 0.239 | 0.302 |
| $\xi_{\mathrm{F}}$ | -2.031 | -2.110 | -2.030 | -2.081 | -2.030 | -2.080 |
| $\xi_{\mathrm{B}}$ | -2.062 | - | -1.997 | - | -1.942 | - |
| $k_{\mathrm{c}}$ $(\mathrm{N}\cdot \mathrm{mm}^{-1})$ | 0.013 | 0.009 | 0.016 | 0.012 | 0.020 | 0.015 |
| $k_{\mathrm{F}}$ $(\mathrm{N}\cdot \mathrm{rad}^{-1})$ | 2.045 | 0.4508 | 3.091 | 1.092 | 4.577 | 1.100 |
| $k_{\mathrm{B}}$ $(\mathrm{N}\cdot \mathrm{rad}^{-1})$ | 1.363 | - | 1.476 | - | 1.488 | - |

Table S3: Mass of the involved origami structures.

| Types | Mass (g) |
|---|---|
| IOS-4 & RIOS-4 | 0.398 |
| IOS-6 & RIOS-6 | 0.795 |
| IOS-8 & RIOS-8 | 1.075 |
| POS | 0.198 |
| Origami hexagram | 0.362 |